\documentclass[conference]{IEEEtran}
\IEEEoverridecommandlockouts
\AtBeginDocument{%
  \providecommand\BibTeX{{%
    \normalfont B\kern-0.5em{\scshape i\kern-0.25em b}\kern-0.8em\TeX}}}

\usepackage[T1]{fontenc}
\usepackage{tgtermes,tgheros,tgcursor}
\usepackage[tbtags]{mathtools}
\mathtoolsset{centercolon}
\usepackage[colorlinks=true, urlcolor=blue, linkcolor=red]{hyperref}
\usepackage{amsfonts}
\usepackage{algorithm}
\usepackage{gnuplottex}
\usepackage{algorithmicx}
\usepackage{algpseudocode}
\algrenewcommand\algorithmicrequire{\textbf{Input:}}
\algrenewcommand\algorithmicensure{\textbf{Output:}}
\usepackage{textcomp}
\usepackage{stfloats}
\usepackage{url}
\usepackage{verbatim}
\usepackage{tcolorbox}
\usepackage{tikz}
\usepackage{comment}
\usetikzlibrary{spy}
\mathchardef\mhyphen="2D
\usepackage{color}
\usepackage{microtype}
\usepackage{multirow}
\usepackage{wrapfig,lipsum}
 \usepackage{booktabs} 
\usepackage{dsfont}
\usepackage{wrapfig}
\usepackage{pgfplots}

\usepgfplotslibrary{fillbetween}
\usepackage{pgfkeys}
\usepackage{lipsum}
\usepackage{mwe}
\usepackage{tikz}
\usetikzlibrary{backgrounds}
\usetikzlibrary{intersections}
\usetikzlibrary{shapes.geometric}
\usetikzlibrary{spy}
\usepackage{pgfplots, pgfplotstable}
\pgfmathdeclarefunction{gauss}{2}{%
  \pgfmathparse{1/(#2*sqrt(2*pi))*exp(-((x-#1)^2)/(2*#2^2))}%
}
\pgfplotsset{compat=1.11,
    /pgfplots/ybar legend/.style={
    /pgfplots/legend image code/.code={%
       \draw[##1,/tikz/.cd,yshift=-0.25em]
        (0cm,0cm) rectangle (3pt,0.8em);},
   },
}
\usepackage{color, colortbl}
\usepackage{xcolor}
\usepgfplotslibrary{colormaps}
\newenvironment{customlegend}[1][]{%
        \begingroup
        \csname pgfplots@init@cleared@structures\endcsname
        \pgfplotsset{#1}%
    }{%
        \csname pgfplots@createlegend\endcsname
        \endgroup
    }%
    \def\addlegendimage{\csname pgfplots@addlegendimage\endcsname}

\pgfplotsset{ 
cycle list={%
{draw=black,mark=star,solid},
{draw=black, mark=square,solid}}}

\usepackage{amsfonts}
\usepackage{textcomp}
\usepackage{breqn}
\usepackage{comment}
 \usepackage{soul}
 
\usepackage{authblk}

\usepackage{caption}
\newcommand{\brokenline}[2][t]{\parbox[#1]{\dimexpr\linewidth-\ALG@thistlm}{\strut\raggedright #2\strut}}

\usepackage{fancyhdr}
\fancypagestyle{firstpage}
{
    \fancyhead[L]{This work has been submitted to the IEEE for possible publication. Copyright may be transferred without notice, after which this version may no longer be accessible.}
    \fancyhead[R]{}
}
\begin{document}


\title{ProAct: Progressive Training for Hybrid Clipped Activation Function to Enhance Resilience of DNNs}
\author[1]{Seyedhamidreza Mousavi}
\author[2]{Mohammad Hasan Ahmadilivani}
\author[2]{Jaan Raik}
\author[2]{Maksim Jenihhin}
\author[1,2]{\\Masoud Daneshtalab}
\affil[1]{Mälardalen University, Västerås, Sweden}
\affil[2]{Tallinn University of Technology, Tallinn, Estonia}
\affil[1]{\{seyedhamidreza.mousavi, masoud.daneshtalab\}@mdu.se}
\affil[2]{\{mohammad.ahmadilivani, jaan.raik, maksim.jenihhin\}@taltech.ee}

\maketitle
\thispagestyle{firstpage}
\begin{abstract}
Deep Neural Networks (DNNs) are extensively employed in safety-critical applications where ensuring hardware reliability is a primary concern. 
To enhance the reliability of DNNs against hardware faults, activation restriction techniques significantly mitigate the fault effects at the DNN structure level, irrespective of accelerator architectures. 
State-of-the-art methods offer either neuron-wise or layer-wise clipping activation functions.
They attempt to determine optimal clipping thresholds using heuristic and learning-based approaches.
Layer-wise clipped activation functions cannot preserve DNNs' resilience at high bit error rates. 
On the other hand, neuron-wise clipping activation functions introduce considerable memory overhead due to the addition of parameters, which increases their vulnerability to faults.
Moreover, the heuristic-based optimization approach demands numerous fault injections during the search process, resulting in time-consuming threshold identification.
On the other hand, learning-based techniques that train thresholds for entire layers concurrently often yield sub-optimal results.

In this work, first, we demonstrate that it is not essential to incorporate neuron-wise activation functions throughout all layers in DNNs.
Then, we propose a hybrid clipped activation function that integrates neuron-wise and layer-wise methods that apply neuron-wise clipping only in the last layer of DNNs.
Additionally, to attain optimal thresholds in the clipping activation function, we introduce ProAct, a progressive training methodology.
This approach iteratively trains the thresholds on a layer-by-layer basis, aiming to obtain optimal threshold values in each layer separately.
Throughout the progressive training, we utilize Knowledge Distillation (KD) to transfer the output class probability from the unbounded activation model (teacher model) to the bounded activation model (student model).
ProAct enhances the fault resilience of DNNs, achieving improvements of up to $6.4$x at high bit error rates.
Moreover, ProAct reduces memory overhead by $91.26$x and $134.28$x compared to the state-of-the-art neuron-wise activation restriction technique, as evaluated on ResNet50 and VGG16 models, respectively. The source codes of the methods experimented in this paper are released at: \url{https://github.com/hamidmousavi0/reliable-relu-toolbox.git}
\end{abstract}

\section{Introduction}
Machine Learning (ML) and, in particular, Deep Neural Networks (DNNs) have recently emerged to play a significant role in 
various applications \cite{sze2017efficient,voulodimos2018deep,lin2020mcunet,loni2022tas}. 
DNN hardware accelerators are widely leveraged in safety-critical applications, e.g., autonomous driving and healthcare, where hardware reliability is a major concern \cite{moolchandani2021accelerating,mittal2020survey,su2023testability,ahmadilivani2024systematic}. 

The reliability of DNN accelerators expresses their ability to produce correct outputs in the presence of hardware faults originating from various phenomena, e.g., soft errors that are caused by colliding high energy particles to digital devices and result in bitflips either in memory or logic \cite{su2023testability,bolchini2022fast}. 
In this regard, resilience pertains to the ability of DNNs to maintain their prediction accuracy in the presence of faults \cite{ahmadilivani2024systematic,ibrahim2020soft,ahmadi2023Deepvigor}.
Due to technology miniaturization, soft error rates have increased in recent years, in particular in SRAM-based memories, leading to significant accuracy drops in DNNs due to corrupted parameters that are stored in memory \cite{su2023testability,azizimazreah2018tolerating,malekzadeh2021impact}.

\begin{figure}[ht!]
    \includegraphics[width=0.45\textwidth]{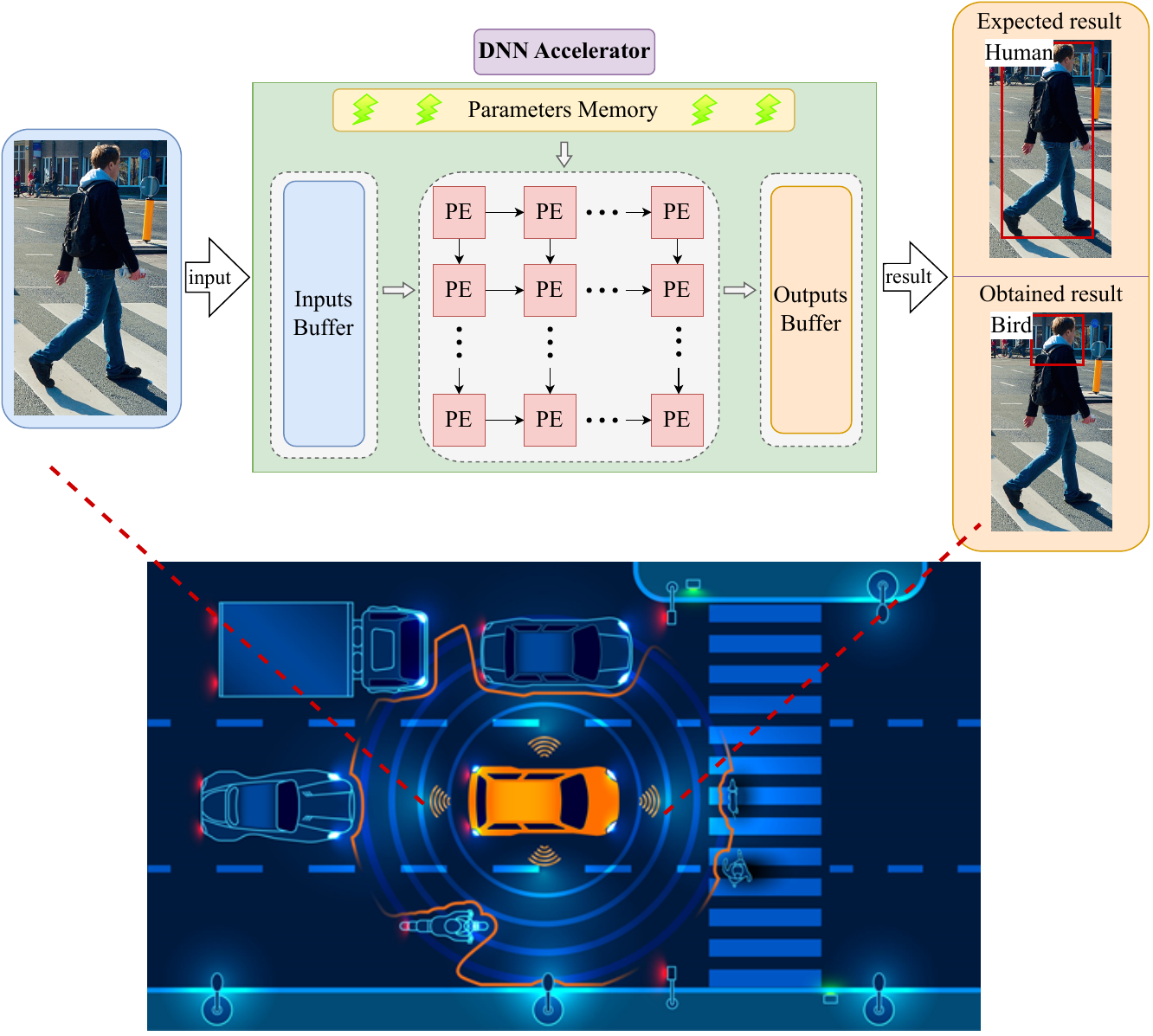}
    \centering
    \caption {Example of the impact of memory faults on the output classification in a safety-critical application.}
    \label{fig:faulty-accel}
\end{figure}

Since DNN accelerators store parameters (i.e., weights and biases) in memory, which is prone to soft errors, the accuracy of DNNs is jeopardized  \cite{neggaz2019cnns,spyrou2022reliability}.
Parameters in memories are not being overwritten as frequently as values in the data path, input buffers, and logic elements (e.g., buffers in PEs), thus, bitflips originating from soft errors in parameters are accumulative and persistent, leading to constantly producing errors throughout the inference of a DNN accelerator.
Fig. \ref{fig:faulty-accel} depicts an example of how soft errors in memories storing DNNs' parameters can lead to a catastrophic outcome for an autonomous vehicle by misclassifying a pedestrian as a bird, which can result in a loss of life.

For the sake of simplicity, without loss of generality, in this paper we are addressing Single Event Upset (SEU) fault effects, i.e. soft errors latched to memory elements. 
SEUs cover a vast majority of Single Event Transient (SET) effects on the logic cells and are significantly more probable than SETs due to the frequent masking of the latter \cite{baumann2005soft}. 
Moreover, in the reliability assessment, we resort to fault injection into parameters of the DNN, while SEU faults may also occur in the neurons and loop counters. 
%
%
These faults, albeit critical, have a marginal impact on the overall reliability assessment result due to the significantly smaller size of the counter registers versus the parameters memory and thereby marginal fault probabilities \cite{lotfi2019resiliency,rech2024artificial}.

Fault-tolerant techniques to enhance the reliability of DNN accelerators against soft errors in memories are carried out at the architecture and algorithm level. Architecture-level techniques are accelerator-specific and exploit hardware redundancy with performance and cost. Whereas they do not apply to the commercial-off-the-shelf and pre-designed IPs. 
However, algorithm-level techniques modify the DNN models in the software executable by any accelerators.
Throughout the literature, several cost-effective algorithm-level fault tolerance techniques for enhancing the reliability of DNNs are presented, such as activation restriction methods \cite{chen2021low,hoang2020ft,ghavami2022fitact}. 

In the activation restriction methods, the rationale is to limit the activation values within layers to mitigate the effect of fault propagation on the output of DNNs mostly produced by large erroneous values. 
To this end, a clipping threshold is obtained for the ReLU activation functions in DNNs, to clip the higher values than the threshold to 0.

In Ranger \cite{chen2021low} and FT-ClipAct \cite{hoang2020ft}, a single threshold is obtained for each layer while fitAct \cite{ghavami2022fitact} assigns a threshold to the ReLU of each neuron throughout a DNN.
To obtain the clipping thresholds, Ranger merely considers them as the maximum activation value in each layer over validation data.
FT-ClipAct presents a heuristic search algorithm to identify the corresponding clipping thresholds.
On the other hand, FitAct proposes a training-based method to obtain the neuron-wise clipped activation function.

Regarding the related research works, the main shortcomings in the previously proposed activation restriction methods are as follows:
\begin{itemize}
    \item The granularity of the existing clipping activation functions is not optimal: 1) Layer-wise method does not effectively prevent errors from propagating to outputs; 2) Neuron-wise method imposes significant overhead on DNNs and increases the number of possible fault locations in the memory.
    \item The existing optimization methods for obtaining clipping thresholds are not only highly time-consuming but also obtain sub-optimal clipping thresholds.
\end{itemize}

In this work, we attempt to address the identified issues in the literature. To the best of our knowledge, for the first time, a novel activation restriction method is introduced that combines layer-wise and neuron-wise clipping incorporated with progressive training employing Knowledge Distillation (KD) \cite{hinton2015distilling,goldblum2020adversarially} to achieve significant resilience with negligible memory overhead in DNNs. 
The contributions of this work are as follows:
\begin{itemize}
    \item Proposing Hybrid Clipped ReLU (HyReLU) activation function restricting the activation values by trainable threshold parameters in a neuron-wise way at the last layer and performs layer-wise in the other layers of DNNs. The proposed HyReLU imposes a negligible memory overhead on DNNs. 
    
    \item Introducing progressive training to obtain trainable clipping thresholds in HyReLU for each layer separately. It transfers the knowledge from baseline DNN models (teacher) to the clipped DNNs by HyReLU (student), leading to more optimal and effective clipping threshold values ensuring high resilience for DNNs.

    \item Comparing the results extensively with the state-of-the-art activation-restriction methods. Results demonstrate that the ProAct method improves the resilience of DNNs up to $6.4$x at high bit error rates, compared to the state-of-the-art methods with a remarkable reduction in memory overhead, i.e., up to 134.28x.
    
    \item Publishing the source codes, not only for the proposed ProAct method but also the implemented state-of-the-art activation restriction methods presented in this paper. To our knowledge, there is no open-source tool in the literature providing activation restriction methods for resilience enhancement of DNNs. The source code and usage instructions for all activation restriction methods, including ProAct, are accessible for researchers on the corresponding GitHub repository\footnote{\url{https://github.com/hamidmousavi0/reliable-relu-toolbox.git}}. 
\end{itemize}

%
Progressive Training for Hybrid Clipped Activation Function Thresholds (ProAct) empowers DNNs to mitigate error propagation to the output to a significantly greater extent and with considerably lower memory overhead than the state-of-the-art methods.


We analyze the distribution of activation values in layer-wise clipped activation functions within fault-free and faulty models to demonstrate that layer-wise clipping suffices for initial layers, while neuron-wise clipping becomes necessary for the final layers. Subsequently, we propose a hybrid neuron-/layer-wise clipped activation function, wherein only the last layer of DNNs employs neuron-wise clipping activation functions, while the preceding layers utilize layer-wise clipping to mitigate memory overhead.
    
We demonstrate that the optimization methods utilized in current state-of-the-art approaches fail to attain optimal threshold values. Subsequently, we introduce a progressive training technique based on knowledge distillation for clipped thresholds, executed layer by layer, to improve the resilience of DNNs.

The remainder of the paper is organized as follows. Section \ref{related-works} overviews the literature on the fault-tolerant techniques for DNNs. Section \ref{preliminaries} presents the preliminaries regarding the activation restriction and knowledge distillation topics. The motivation for the research is presented in Section \ref{Motivation}. The ProAct method is described in Section \ref{method}. The experimental setup and results are presented in Section \ref{experiments}. Finally, the paper is concluded in Section \ref{conclusion}. 
\section{Related Works} \label{related-works}

In this section, the research works attempting to improve the reliability of DNNs are overviewed. Techniques for improving the reliability of DNN accelerators against soft errors in memories can be considered at two levels of system abstraction:
\begin{itemize}
    \item Architecture level: The computing units or memory components of the DNN accelerators are either designed to be reliable (i.e., hardened) or redundant units are included \cite{liu2021hyca,li2020soft,hong2022statistical,ahmadilivani2023enhancing}.
    \item Algorithm level: The structure of the DNN model is manipulated or fault-aware training is performed to improve the resiliency of the DNN \cite{cavagnero2022fault,zahid2020fat,santos2019analyzing,ghavami2022fitact}. 
\end{itemize}

Architecture-level techniques are designed specifically for concrete DNN accelerators, and they apply hardware redundancy leading to performance, area, and power overheads.
However, algorithm-level techniques are concerned with the DNN model itself and can be applied to any accelerator \cite{zhan2021improving,schorn2019efficient}.
Throughout the literature, the effective algorithm-level techniques for improving the reliability of DNNs are presented as follows:
\begin{itemize}
    \item Fault correction: Faults are corrected using Error Correction Codes (ECC) \mbox{\cite{jang2021mate,lee2022value}}, or Algorithm-Based Fault Tolerance (ABFT) \mbox{\cite{zhao2020ftcnn}} approaches.
    \item Inherent resilience improvement: Different approaches to improve the inherent resilience of DNNs including quantization and outlier regularization \mbox{\cite{ozen2021snr}}, fault-aware training \mbox{\cite{zahid2020fat,cavagnero2022transient}}, and activation restriction \mbox{\cite{ghavami2022fitact,zhan2021improving,ali2020erdnn,chen2021low,hoang2020ft}}.
\end{itemize}

ECC and ABFT methods are conventional and well-established fault tolerance methods to detect and correct faults using numerical and computational overheads.
However, the efficacy of these techniques in fault correction highly depends on the overhead they introduce, whereas the state-of-the-art approaches tend to improve the inherent resilience of DNNs in more effective and efficient ways \mbox{\cite{zahid2020fat,ghavami2022fitact,ozen2021snr}}.
Nonetheless, the mentioned approaches for DNNs' inherent resilience improvement are orthogonal to one another and can be applied as a combination to enhance their inherent resilience.  

The process of quantization and outlier regularization offers the potential to restrict the numerical range within a DNN, thereby eliminating the possibility of generating excessively large values due to faults.
Nonetheless, quantization necessitates the utilization of hardware accelerators specifically designed to handle the operations associated with the respective data types.
Quantization can be deployed on general-purpose devices as well, however, these carry out the floating-point arithmetic leading to the reliability issues of floating-point data types. 

In the research works leveraging fault-aware training methods, all the DNN's parameters are retrained while fault injection is being performed. 
Fault-aware training for stuck-at and transient faults at activations are presented in \cite{cavagnero2022fault} and \cite{zahid2020fat}, respectively. 
It is shown, including by radiation experiments, that fault-aware training methods effectively improve the resilience of DNNs \cite{gambardella2022accelerated,maillard2022radiation}.
However, these methods retrain the entire DNN which requires updating all the model parameters. This is excessively complex and compute-expensive and requires the possibility of retraining for a pre-trained DNN.

As mentioned, activation restriction methods attempt to restrain the activation values within layers to alleviate the effect of fault propagation on the DNNs outputs, produced by large erroneous activation values. 
Assuming faults are either in parameters or computations, neurons' erroneous output activations can be detected and handled after the generation of their respective outputs using activation restriction methods \cite{hoang2020ft}.
These methods do not require retraining the entire model and do not suffer from the complexity of fault-aware training. In addition, they are non-intrusive in the sense that they do not require any changes to an accelerator. 

Piece-wise Rectified Linear Unit (ReLU) is proposed in \cite{ali2020erdnn} that finds thresholds to split ReLU into different ranges by training and applies predefined coefficients to its outputs. However, the effect of large faulty activation values remains. 
Zhan et al. \cite{zhan2021improving} have proposed another method to find the thresholds of ReLU in a layer called BReLU and clip the output to the threshold value itself. BReLU maps the faulty activations in a layer to a non-zero value within that layer, while DNNs are shown to be more resilient if faulty values are replaced with a value near 0.

Ranger \cite{chen2021low} presents a clipped ReLU that bounds the layer activations' output to 0 in case their value is higher than a fixed threshold. 
The threshold values are obtained from the maximum values at each layer seen in a validation set of the dataset. 
The method of clipping out-bound values to 0 in Ranger is demonstrated to be effective, however, the method does not provide optimal clipping thresholds.  

Hoang et al. have analyzed various boundary values on the model's accuracy \cite{hoang2020ft}.
Their method, named FT-ClipAct, attempts to find optimal boundary values for each layer that are not necessarily the maximum values of the layers' activations and are smaller than the maximum bounds.
The authors propose a heuristic interval search algorithm based on the fault injection process to find appropriate threshold values for the ReLU activation function at each layer. 
FT-ClipAct incurs significant computational overhead in determining the thresholds for DNNs' layers due to the injection of faults at each step of the search algorithm.
Therefore, it is unfeasible to employ it for every single neuron in a DNN. 

FitAct \cite{ghavami2022fitact} proposes an activation function based on the \textit{sigmoid} function that is differentiable to boundary values to optimize them with a gradient-based algorithm. 
FitAct considers the boundary values for each neuron and smoothly maps the activation outputs to 0.
Furthermore, Fitact demonstrates that for fixed-point representation, the optimal threshold values that maintain the baseline accuracy of the fault-free model tend to be smaller.  
While FitAct effectively enhances the resilience of DNNs, it concurrently elevates both memory overhead and the likelihood of faults occurring in the activation functions' parameters. 
This indicates that as the number of parameters in DNNs grows, the likelihood of faults occurring in the threshold values also increases, thereby diminishing the resilience. 
Furthermore, FitAct trains all the threshold parameters in the clipping activation function simultaneously, which decreases the possibility of providing the optimal threshold for each activation function.

Therefore, there is a need for new activation restriction methods in which more optimal thresholds can be obtained and less memory overhead is incurred to DNNs. This paper attempts to address these shortcomings in the literature. 

\section{Preliminaries} \label{preliminaries}
\label{sec:Preliminaries}
\subsection{Clipping Activation Functions}
Deep Neural Networks (DNNs) that are exploited for image classification mainly consist of two main types of computational layers: convolution and the fully-connected.
An activation function follows these layers to take non-linearity into account. 
ReLU is the most frequently used activation function in DNNs which is defined as follows:
\begin{equation}
    ReLU(x) = \max (0,x)
\end{equation}

ReLU has a remarkable impact on DNN training in terms of efficiency and accuracy. However, it passes all positive values leading to resilience issues once a fault produces large values in the activations.
This phenomenon decreases the classification accuracy of the model significantly \cite{hoang2020ft}.
Therefore, it is possible to create a clipped ReLU activation function to increase the resilience of DNNs.

The primary strategy for restricting ReLU's output values is to use threshold values for each layer to prevent crossing the large values.
To achieve this, a clipped version of the ReLU activation function for each layer is proposed in \cite{chen2021low,hoang2020ft}:
\begin{equation}
  ReLU_{clipped}(x) =
    \begin{cases}
      x & \text{if $0 \leq x \leq \lambda $}\\
      0 & \text{otherwise}
    \end{cases}       
\end{equation}
where $\lambda$ is the clipping threshold and any value above this threshold is considered faulty, and its value is clipped to 0.

As mentioned, Ranger \cite{chen2021low} obtains the threshold values ($\lambda$) based on the maximum value in the activation of each layer on validation data. FT-ClipAct \cite{hoang2020ft} finds the thresholds for each layer by performing an interval search algorithm, which results in a lower accuracy drop than Ranger in the presence of faults. FitAct \cite{ghavami2022fitact} introduces a neuron-wise smooth activation function and is used as a gradient-based optimization method to obtain all thresholds efficiently and provides better results than FT-ClipAct in terms of accuracy drop, with considerably higher memory overhead.

\subsection{Knowledge Distillation} 
Knowledge Distillation (KD) is a teacher-student paradigm, where the student model learns to mimic the output class probability of the teacher model \cite{wang2021knowledge}. 
The most common KD technique for mimicking the output of the teacher model is known as \textit{soft targets} that is based on the output logits (activation outputs before \textit{softmax}) of teacher and student models \cite{hinton2015distilling}.
The smooth, extended version of the \textit{softmax} function transforms the logits of a neural network into soft probabilities $P$. The formulation is as follows:
\begin{equation}\label{eq3}
    P(z_i,T)= \frac{\exp(z_i/T)}{\sum_{j}\exp(z_j/T)}
\end{equation}
where $z$ is the logits, $i$ is the top output class and $j$ refers to all output classes, and $T$ indicates is a hyper-parameter called temperature.

KD can be expressed as a minimization problem that minimizes an expected ($\mathbb{E}_{X \sim D}$) function based on soft targets, as:
\begin{equation}\label{eq4}
    \min_{\theta} \mathbb{E}_{X \sim D} [KL(P(f_{s}(X,\theta),T)|| P(f_{t}(X),T))]
\end{equation}
where $f_{s}(\cdot,\theta)$ and $f_{t}(\cdot)$  are the logit outputs of the student model ($s$) with parameters $\theta$ and pretrained teacher model ($t$), respectively. $KL$ shows the KL-divergence distance which measures the difference between two distributions, and $X$ is an input to the model that is drawn from the data distribution $D$.

The main objective of knowledge distillation is to reduce the gap between the outputs of the student and teacher models in the last layers.
By optimizing Eq.~\eqref{eq4}, the logit outputs of the student model eventually coincide with those of the teacher model.

\section{Research Motivation} \label{Motivation}
\label{sec:Motivation}
The existing activation restriction methods in the literature incur a notable accuracy drop at high Bit Error Rates (BERs) and memory overhead accompanied by a time-consuming and computationally expensive process to obtain the clipping thresholds. 
The primary reason for the mentioned shortcomings stems from the optimization process to determine the clipping thresholds for clipping activation functions within DNN layers or neurons.

It is stated that the optimal clipping threshold value for an activation function (layer-wise or neuron-wise) is the minimum possible value that maintains the accuracy of the fault-free DNN model ~\cite{hoang2020ft,ghavami2022fitact}.
The heuristic optimization method in FT-ClipAct ~\cite{hoang2020ft} demands extensive computation overhead as it involves conducting fault injections at every step of the search process and finding sub-optimal thresholds due to the limited number of search steps.

The gradient-based optimization method in FitAct \cite{ghavami2022fitact} improves the resilience of DNNs compared to FT-ClipAct as well as reduces its computational overhead. However, the obtained thresholds for neurons are not optimal since all of them are trained simultaneously in each backward pass. As DNNs possess numerous neurons, this optimization process may not necessarily lead to an optimal local minimum for the clipping thresholds of activation functions for all neurons of a DNN. 

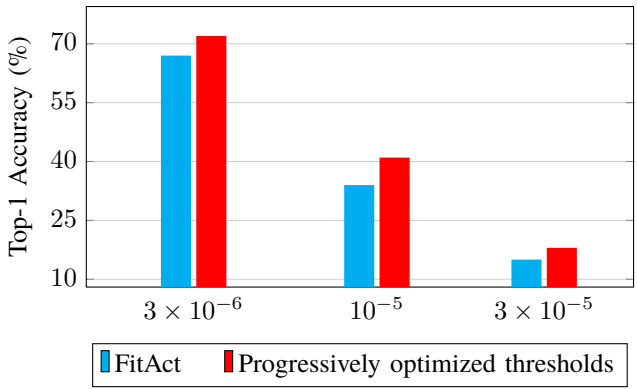
\begin{figure}
    \centering
    \begin{tikzpicture}
    \pgfplotsset{
      log x ticks with fixed point/.style={
          xticklabel={
            \pgfkeys{/pgf/fpu=true}
            \pgfmathparse{exp(\tick)}%
            \pgfmathprintnumber[fixed relative, precision=3]{\pgfmathresult}
            \pgfkeys{/pgf/fpu=false}
          }
      },
      log y ticks with fixed point/.style={
          yticklabel={
            \pgfkeys{/pgf/fpu=true}
            \pgfmathparse{exp(\tick)}%
            \pgfmathprintnumber[fixed relative, precision=3]{\pgfmathresult}
            \pgfkeys{/pgf/fpu=false}
          }
      }
    }
  \centering
  \begin{axis}[xmode=log,log basis x={2},
        width=\columnwidth,
        height=0.6\columnwidth,
        ybar  ,
        bar width=0.4cm,
        major grid style={draw=white},
        grid=minor,
        minor ytick={0.1,0.25,0.4,0.55,0.7},
        grid style={line width=.1pt, draw=gray!40},
        major grid style={line width=.2pt,draw=gray!50},
        enlarge y limits={value=.1,upper},
        ymin=0.08, ymax=0.73,
        xmin =0.000002, xmax=0.00004,
        tickwidth=0pt,
        enlarge x limits=true,
        legend style={
            at={(0.5,-0.2)},
            anchor=north,
            legend columns=-1,
            /tikz/every even column/.append style={column sep=0.5cm}
        },
       ylabel={Top-1 Accuracy (\%)},
       legend style={draw=black, text opacity = 1,row sep=0pt, font=\selectfont},
       xtick={ 0.000003,0.00001,0.00003}, 
       ytick={0.1,0.25,0.4,0.55,0.7},
        xticklabels = {\strut $3\times10^{-6}$,\strut $10^{-5}$,\strut  $ 3\times10^{-5}$},
        yticklabels = {\strut $10$,\strut $25$,\strut $40$, \strut $55$, \strut $70$ },
    ]
    \addplot [draw=none, fill=cyan] coordinates {(0.000003,0.67)(0.00001,0.34)(0.00003,0.15)};
    \addplot [draw=none, fill=red] coordinates {(0.000003,0.72)(0.00001,0.41)(0.00003,0.18)};
   \legend{FitAct, Progressively optimized thresholds} 
  \end{axis}
  \end{tikzpicture}
    \caption{Top1-Accuracy of AlexNet under different BERs employing FitAct and progressively optimized thresholds.}
    \label{fig:acc-drop}
\end{figure}

To illustrate the aforementioned shortcoming, we calculate the clipping threshold values for activation functions in the AlexNet using FitAct. 
Afterwards, we attempt to progressively reduce the clipping threshold values for the neurons layer by layer while ensuring that the model's baseline accuracy remains unaffected.
We accomplish this by training the threshold parameters for each layer individually for 5 epochs, using a higher weight decay hyper-parameter.
Fig. \ref{fig:acc-drop} illustrates the results for AlexNet resilience based on its accuracy under different BERs into parameters, after minimizing the neurons' thresholds in each layer progressively. 

It is observed that the obtained clipping thresholds by FitAct are not the optimal values and it is possible to identify more appropriate clipping threshold values to improve the resilience of DNNs. 
These results demonstrate the necessity for a new training mechanism to optimize clipping threshold values. In this paper, we introduce the ProAct algorithm, which progressively trains threshold values layer by layer accompanied by Knowledge Distillation (KD).

\begin{figure}

    \centering
    \includegraphics[width=\columnwidth, height=\columnwidth]{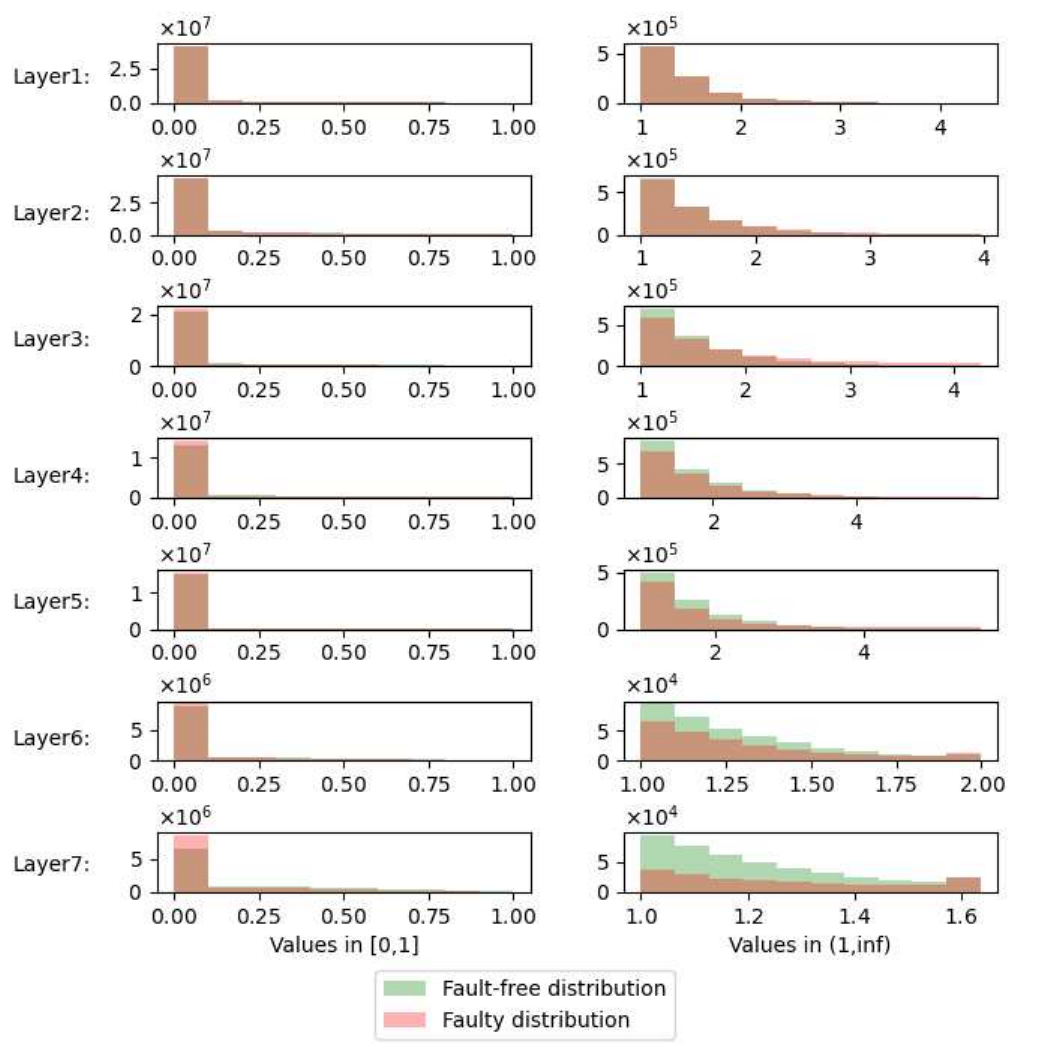}
    \caption{The distribution of output activation values for the AlexNet model on the CIFAR-10 dataset after applying the FitAct algorithm to find threshold parameters.}
    \label{fig:enter-label}
\end{figure}

Moreover, the main factor contributing to memory overhead in FitAct is the implementation of clipped thresholds at the level of individual neurons. 
Applying an individual clipping threshold to each neuron not only enlarges memory overhead but also increases the probability of memory faults as there are more stored parameters. 
To comprehend the impact of neuron-wise activation restriction, we examine error propagation through \textit{FitAct}-instrumented AlexNet by illustrating the distribution of activations of each layer without and with faults into parameters (BER = $3e-5$) in Fig. \ref{fig:enter-label}.
To enhance the visualization, the distribution of values is partitioned into two ranges, the left-hand side column presents the activation values between [$0,1$] and the right-hand side one shows them in $(1,\infty)$, respectively.

The distribution of activations in both fault-free and faulty models exhibits a similarity in the initial layers. 
While, by proceeding through the depth of the model, the disparity between the distribution of activations in fault-free and fault models becomes more pronounced.
This phenomenon reveals that the errors are mostly amplified in the last layers of DNNs where it is crucial to harness them.
This observation suggests that the output activations in the initial layers of DNNs can be restricted by layer-wise clipping thresholds and the last layer can be restrained by neuron-wise ones.
As a solution, we introduce a hybrid clipped activation function that incorporates neuron-wise thresholds specifically for the last layer of DNNs and layer-wise thresholds for the rest of the layers, aiming to decrease memory overhead and enhance resilience.

\section{Methodology} \label{method}
\label{sec:Methodology}
Building upon the insights from the previous section, we introduce ProAct, a progressive training approach for clipping activation function thresholds, implemented in a hybrid manner: neuron-wise exclusively in the last layer of DNNs and layer-wise in all other layers.
Progressive training aims to minimize clipping thresholds for activation functions and enhance DNNs resilience while maintaining their baseline accuracy.
Moreover, the hybrid clipped activation function mitigates memory overhead and reduces the occurrence of faults in memory locations.

\subsection{\textbf{Hybrid Clipped ReLU and Its Memory Overhead}}
\label{sec:Methodology:P-Act}
To propose a hybrid clipped ReLU activation function, it is essential to incorporate neuron-specific thresholds for the neurons in the final layer of DNNs, while employing layer-specific thresholds for the neurons in the preceding layers.
In addition, to utilize the gradient-based optimization method, it is necessary to create a differentiable version of the activation function. 
To achieve these objectives, we introduce a hybrid clipped ReLU activation function (HyReLU), which draws inspiration from the sigmoid activation function ($\sigma$), in Eq.~\eqref{eq:hyrely}.
This function guarantees smooth transitions around the threshold values ($\lambda$) utilized in the clipped ReLU.  


\begin{equation}
HyReLU(x,\lambda,l) = 
    \begin{cases} 
    \max\{0, x_i [1-\sigma(k[\lambda_i-x_i])]\} \; if \;l=L
            \\
    \max\{0, x [1-\sigma(k[\lambda-x])]\}\;\; otherwise
        \end{cases}
    \label{eq:hyrely}
\end{equation}
In Eq.~\eqref{eq:hyrely}, \textit{x} is an input activation, $\lambda$ is a trainable parameter representing the value of the clipping threshold in the respective neuron/layer and $k$ is a hyper-parameter determining the slope for the smooth transition to 0, which is obtained through cross-validation. $L$ indicates the last layer index.
This equation expresses that the values larger than $\lambda$ in a layer are considered erroneous and are smoothly clipped to 0. 

HyReLU is employed across all neurons of the last layer of DNNs (i.e., the layer preceding the output layer), with each having a distinct trained $\lambda_i$. For other layers, the function is applied separately, with each layer possessing its own trained $\lambda$. Consequently, the memory overhead introduced to a DNN with the HyReLU is formulated in Eq.~\eqref{eq:mem-overhead}.

\begin{equation}
    Memory\;Overhead\;=\frac{\#Layers\;+\;\#Neurons_{last\;layer}}{\#Parameters_{DNN}}
    \label{eq:mem-overhead}
\end{equation}


\subsection{\textbf{ProAct: Progressive Training for HyReLU Activation Function}}
\label{sec:Methodology:Knowledge Distillation}
\begin{figure*}[ht!]
	\begin{center}
		\includegraphics[width=\textwidth]{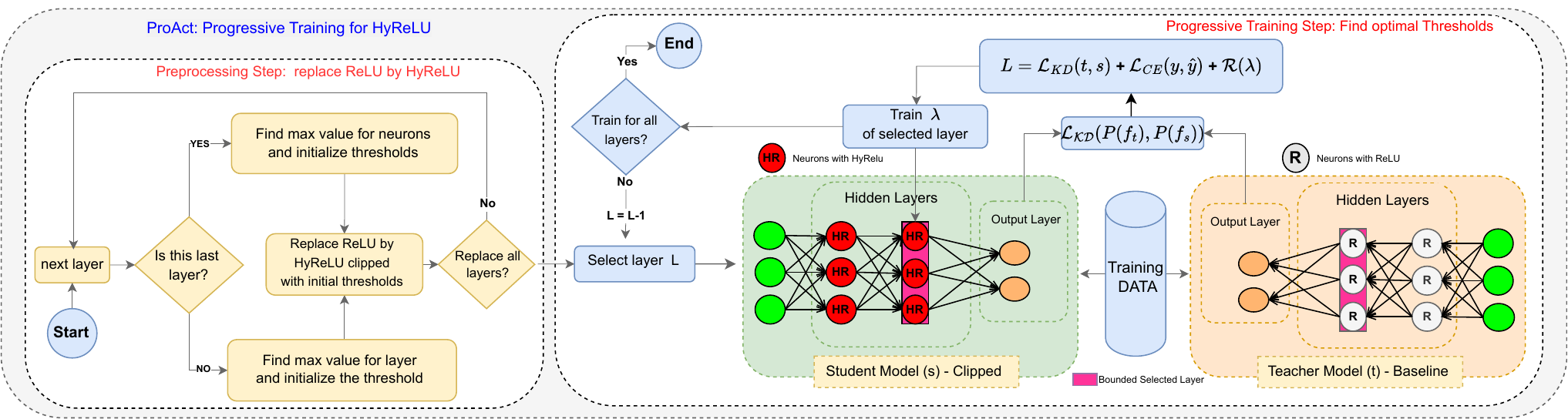}
	\end{center}	
\caption{Hybrid Progressive training based on Knowledge Distillation }
\label{fig:training-steps}
\end{figure*}

%
%
To obtain the best clipping thresholds ($\lambda$) in HyReLU for each layer/neuron in a DNN, we propose ProAct, a layer-wise progressive training method exploiting knowledge distillation. 
Figure~\ref{fig:training-steps} depicts an outline of the ProAct approach, where the purpose is to find an optimal $\lambda$ for each HyReLU without breaching the maximum permitted accuracy drop and memory overhead.

ProAct trains the clipping threshold of each layer separately, from the last to the first layer. 
ProAct includes two main steps to find the threshold parameters 1) Preprocessing and 2) Progressive training. 
In the first step, we start by profiling the model on validation data to determine the initial values for the threshold parameters.
Specifically, we initialize the threshold parameters with the maximum activation value observed by the corresponding layer/neuron on the validation dataset.
Then, we replace all ReLU activation functions with the proposed HyReLU, using the initial threshold parameters (preprocessing step in Algorithm~\ref{method:alg:FKD} (1-8 lines)). 
%
%
Within the progressive training step, we progressively select the layers from the last layer to the first one and train the threshold parameters ($\lambda$s) in the HyReLU of the selected layer through the KD-based training.
The clipping threshold of the target layer is trained using KD and this process continues down to the first layer (Training step in Algorithm~\ref{method:alg:FKD} (9-15 lines)).

The proposed training method utilizes KD in a way that the clipping thresholds in HyReLU are trained based on the supervision of the unbounded fault-free (baseline) model.
Through this process, the purpose is to mimic the output values of the unbounded fault-free model in the modified model with the HyReLU activation function.
The pre-trained baseline model including ReLU is used as the teacher model that includes the golden output values and the modified DNN is the student model that has the same structure as the teacher model, but ReLU replaced by HyReLU.

The loss function $\mathcal{L}_{KD}$ is computed based on the Kullback–Leibler divergence (KL) distance \cite{shlens2014notes} between the distribution of output values in the student and the teacher model as:
\begin{dmath}
    \mathcal{L}_{KD}(X,s,t) = \mathbb{E}_{x\sim D} KL\bigg(P(f_{s}(x,T,\lambda))||P(f_{t}(x),T)\bigg) 
\end{dmath}
where $P(f(.,T))$ show the soft output logits with temperature parameter $T$.

Therefore, the whole loss function ($L$) that we use to train the threshold parameters in the selected layer is: 
\begin{dmath}\label{eq8}
    \mathcal{L}(X,Y,\lambda) = \mathcal{L}_{KD}(X,s,t) + \mathbb{E}_{x,y\sim D} L_{ce}(f_{s}(x,\lambda),y) + \gamma R(\lambda)
\end{dmath}
where the $L_{ce}$ and $R(\lambda)$ show the cross entropy loss function and $l_2$-regularization.
$l_2$-regularization helps to constrain the magnitude of the threshold parameters, preventing them from becoming excessively large.
%

%
\begin{algorithm}[hbpt]
\caption{ProAct: Progressive Training for HyReLU Activation Function  }
\label{method:alg:FKD}
\begin{algorithmic}[1]
\Require{The unbounded teacher and bounded student models ($t,s$), learning rate ($\alpha$), Regularization parameter ($\gamma$), number of epochs ($N$), BERs = $[10^{-6},3 \times 10^{-6},10^{-5},3\times 10^{-5},10^{-4}]$};
\Ensure{Resilience DNN;}
    \Statex \textit{\textbf{Preprocessing Step }}
    \For{$l \in [1,2,\cdots,L]$}:
    \If {$l = L $}:
    \State {Profile the model to find max value in each \textit{\textbf{neuron}};}
    \Else :
    \State {Profile the model to find max value in each \textit{\textbf{Layer}};}
    \EndIf;
    \State \brokenline{Initial the threshold parameters ($\lambda$) based on max values;}
    \EndFor
    \Statex \textit{\textbf{Progressive Training Step}}
    \For{$l \in [L,L-1,\cdots,1]$}:
    \For{$i \gets 1$ to $N$}: 
    \State \brokenline{Compute $\mathcal{L}$ (loss function) based on Eq.~\eqref{eq8};}
    \State \brokenline{Compute  $\partial\mathcal{L} / \partial\lambda$  where $\lambda$ : $\nabla_{\lambda}\mathcal{L}(X,Y,\lambda)$;}
    \State \brokenline{Update $\lambda$ via Adam optimizer;}
    \EndFor;
    \EndFor;
\end{algorithmic}
\end{algorithm}

\section{Experiments} \label{experiments}
\label{sec:Experiments}
\subsection{Experimental Setup}  \label{sec:Experiments:setup}
The proposed method ProAct is applied to and evaluated on three DNNs: AlexNet \cite{krizhevsky2017imagenet}, VGG-16 \cite{simonyan2014very}, and ResNet-50 \cite{he2016deep}, all trained on both CIFAR-10 and CIFAR-100 datasets.
Their baseline classification accuracy on the test sets is shown in Table~\ref{table:params-count}. 
It is noteworthy that experiments in this work exploit 32-bit fixed point data type representation in which the Most Significant Bit (MSB) is for sign, 15 bits are for the integer and 16 bits are for the fraction. 
All experiments in this paper are performed on Nvidia\textsuperscript{®} A4000-16GB GPU.

\begin{table}[t]
\centering
\caption{Baseline accuracy for each baseline CNNs}
\label{tab:numberparam}
\begin{tabular}{cccc}
\hline
\textbf{Networks} & AlexNet & VGG-16 & ReNet-50 \\ 
\hline
\hline
\textbf{Accuracy for CIFAR-10} & 81.67\% & 89.87\%  & 91.11\% \\
\hline
\textbf{Accuracy for CIFAR-100} & 55.44\% & 65.45\%  & 74.37\% \\
\hline
\hline
\end{tabular}
\label{table:params-count}
\end{table}
To demonstrate the excellence and effectiveness of ProAct with respect to the state-of-the-art, results are compared to Ranger \cite{chen2021low}, FT-ClipAct \cite{hoang2020ft}, and FitAct \cite{ghavami2022fitact} methods.
We implement Ranger in, both, layer-wise and neuron-wise manners and use a random small part of training data (3000 out of 60000 in both CIFAR-10 and CIFAR-100 datasets) as the validation data to find the maximum values for the layers/neurons.
FT-ClipAct is implemented layer-wise and FitAct is implemented neuron-wise, as they are presented in \cite{hoang2020ft} and \cite{ghavami2022fitact}, respectively.  
%

To obtain a quantitative comparison with the existing works, we carry out the reliability assessment by injecting random bit-flip faults into the parameters of CNNs, including weights, bias, and parameters of clipping activation functions as the fault space. 
Bits are randomly selected and flipped based on the 32-bit fixed-point data representation. We consider different Bit Error Rates (BERs) to flip multiple bits to model the accumulative effect of faults into the memory through time. 
The experimented BERs are $[10^{-7}, 3\times10^{-7},10^{-6},3 \times 10^{-6},10^{-5}, 3\times 10^{-5}]$.

Fault injection experiments are repeated 500 times for each BER and average results for Top-1 accuracy are reported and compared. For the fault injection, we adopt and extend PyTorchFI \cite{mahmoud2020pytorchfi} to consider clipping thresholds in the fault space which is developed on top of PyTorch. 
The training iterations consist of 50 epochs for neuron-specific clipping thresholds in the final layer and 20 epochs for layer-wise HyReLUs. This ensures comparable computational overhead to FitAct, which employs 150 epochs. We initialize the learning rate at 0.01, halving it every 10 epochs, and utilize a batch size of 128.
\subsection{Experimental Results}

\subsubsection{\textbf{Effect of Activation Restriction Methods on DNNs' Baseline Accuracy- and Memory Footprint}}

As mentioned, the clipping thresholds in any activation restriction method are obtained through validation data (not test data).
On the other hand, the main requirement of applying them is that they are required not to drop the baseline accuracy of fault-free DNNs over unseen test data. 
Table~\ref{tab:baseline-acc-drop} shows the impact of activation restriction methods on the baseline accuracy for each DNN after application. It is observed that: 
\begin{itemize}
    \item Ranger has the least effect on the accuracy drop compared to the other methods. Since Ranger considers the clipping threshold as the maximum value seen in validation data (either for neurons or layers), the obtained clipping thresholds are large enough not to affect the inference of the test data. As a result, in fault-free DNNs in Table~\ref{tab:baseline-acc-drop} Ranger reduces the accuracy by less than 0.2\%. However, this method does not improve DNNs' resilience as effectively as other methods, as shown in the next subsection.   
    
    \item FT-ClipAct introduces the largest accuracy drop through all methods, from 0.9\% in AlexNet trained up to 4.68\% in ResNet-50 both trained on CIFAR-10. Such accuracy drop is significant for DNNs, especially in safety-critical applications, and decreases the effectiveness of the applied activation restriction method. Since the heuristic search algorithm is very complex and slow, it is exploited with a small subset of the training data (1000 out of 60000 in both CIFAR-10 and CIFAR-100). Therefore, the obtained thresholds are not optimal and influence the accuracy considerably. 
    
    \item The accuracy drop induced by applying ProAct is always less than 1\% which is negligible. Moreover, in all cases, ProAct reduces the baseline accuracy of DNNs less than both FT-clipAct and FitAct methods. This is due to the progressive training method, which ensures that the optimal threshold for each layer is found separately without sacrificing accuracy.
\end{itemize}

\begin{table}[t]
\caption{Baseline Accuracy drop of DNNs by different activation function restriction methods.}
\resizebox{\columnwidth}{!}{%
\begin{tabular}{cccccc}
\hline
\begin{tabular}[c]{@{}c@{}}Activation \\ restriction \\ method\end{tabular}  & \begin{tabular}[c]{@{}c@{}}Ranger \\ (layer-wise)\end{tabular}  & \begin{tabular}[c]{@{}c@{}}Ranger \\ (neuron-wise)\end{tabular} & \begin{tabular}[c]{@{}c@{}}FT-ClipAct\\ (layer-wise)\end{tabular} & \begin{tabular}[c]{@{}c@{}} FitAct \\ (neuron-wise)  \end{tabular} & \begin{tabular}[c]{@{}c@{}} ProAct \\ (hybrid)  \end{tabular} \\ \hline \hline
\begin{tabular}[c]{@{}c@{}}AlexNet \\ CIFAR-10\end{tabular} & 0.00\% & 0.00\% &0.90\% & 0.78\% & 0.29\% \\ \hline
\begin{tabular}[c]{@{}c@{}}AlexNet \\ CIFAR-100\end{tabular} & 0.01\% &  0.03\% &2.40\% & 0.64\% & 0.51\% \\ \hline
\begin{tabular}[c]{@{}c@{}}VGG-16 \\ CIFAR-10\end{tabular} & 0.00\% & 0.02\% &1.15\% & 1.14\% & 1.00\% \\ \hline
\begin{tabular}[c]{@{}c@{}}VGG-16 \\ CIFAR-100\end{tabular} & 0.00\% & 0.07\% &1.69\% & 0.83\% & 0.35\% \\ \hline
\begin{tabular}[c]{@{}c@{}}ResNet-50 \\ CIFAR-10\end{tabular} & 0.15\% & 0.19\% &4.69\% & 0.30\% & 0.22\% \\ \hline
\begin{tabular}[c]{@{}c@{}}ResNet-50 \\ CIFAR-100\end{tabular} & 0.00\% & 0.08\% &1.60\% & 0.03\% & 0.10\% \\ \hline
\end{tabular}
}

\label{tab:baseline-acc-drop}
\end{table}

The existing methods are either neuron-wise or layer-wise, which lay different memory overhead on DNNs.
Layer-wise approaches introduce new clipping threshold parameters to DNNs proportional to the number of layers which is a negligible overhead. 
Whereas neuron-wise approaches lay a remarkable overhead as the number of neurons in the DNN is huge, leading to an increase in fault locations within the memory.
However, the ProAct memory footprint is limited since it is a hybrid neuron-wise and layer-wise activation function.

Table~\ref{tab:mem-overhead} compares the memory overhead of neuron-wise, layer-wise, and the proposed hybrid approach for activation restriction methods. It is observed that ProAct significantly reduces memory overhead compared to neuron-wise techniques such as FitAct, ranging from $10.5$x to $134.28$x, while still ensuring enhanced accuracy in protecting DNNs against faults.

\begin{table}[t]
\caption{Comparison of memory overhead for neuron-wise, layer-wise, and hybrid activation restriction methods}
\centering
\begin{tabular}{cccc}
\hline
\begin{tabular}[c]{@{}c@{}}Activation \\ restriction \\ method\end{tabular}  & neuron-wise & layer-wise & \textbf{Hybrid (ProAct)} \\ \hline \hline
\begin{tabular}[c]{@{}c@{}}AlexNet \\ CIFAR-10\\ \end{tabular} &0.29&3$\times 10^{-5}$ &\textbf{0.017} \\ \hline
\begin{tabular}[c]{@{}c@{}}AlexNet \\ CIFAR-100\\ \end{tabular} &0.21 &3.5$\times 10^{-5}$ &\textbf{0.020}   \\ \hline
\begin{tabular}[c]{@{}c@{}}VGG-16  \\ CIFAR-10\\ \end{tabular} &1.88 &8.84$\times 10^{-5}$ & \textbf{0.014}  \\ \hline
\begin{tabular}[c]{@{}c@{}}VGG-16  \\ CIFAR-100\\ \end{tabular} &0.85& 4.46$\times 10^{-5}$ & \textbf{0.012} \\ \hline
\begin{tabular}[c]{@{}c@{}}ResNet-50  \\ CIFAR-10\\ \end{tabular} &12.23&2$\times 10^{-4}$&\textbf{0.134} \\ \hline
\begin{tabular}[c]{@{}c@{}}ResNet-50  \\ CIFAR-100\\ \end{tabular} &12.23&2$\times 10^{-4}$&\textbf{0.134} \\ \hline
\end{tabular}

\label{tab:mem-overhead}
\end{table}

\begin{table}[b!]
\caption{Comparing the accuracy drop of DNNs using different activation restriction methods under fault injection.}
\resizebox{\columnwidth}{!}{%
\begin{tabular}{ccccc}
\hline
DNNs                & BER  & FT-ClipAct & FitAct & \textbf{ProAct} \\ \hline \hline
AlexNet CIFAR-10    & 1E-6 & 7.67\%     & 4.34\% & \textbf{2.52\%}  \\ \hline
VGG-16 CIFAR-10     & 3E-6 & 3.19\%     & 2.61\% & \textbf{1.88\%}  \\ \hline
ResNet-50 CIFAR-10  & 1E-6 & 9.09\%     & 1.53\% & \textbf{1.42\%}  \\ \hline
AlexNet CIFAR-100   & 3E-7 & 7.89\%     & 6.31\% & \textbf{5.28\%} \\ \hline
VGG-16 CIFAR-100    & 1E-7 & 6.74\%     & 6.34\% & \textbf{4.40\%}  \\ \hline
ResNet-50 CIFAR-100 & 3E-7 & 12.75\%    &  11.24\% & \textbf{9.37\%}  \\ \hline
\end{tabular}%
}
\label{tab:acc-drop-ber}
\end{table}

\subsubsection{\textbf{Resilience Analysis of Activation Restriction Methods Using Fault Injection}}
 Fig.~\ref{fig:experiments:acc:cifar10} and Fig.~\ref{fig:experiments:acc} depict the Top-1 accuracy of DNNs leveraging different activation restriction methods on CIFAR-10 and CIFAR-100 respectively, under fault injection campaigns as described in Subsection \ref{sec:Experiments:setup}.
The right column in both figures magnifies the results to highlight the impact of ProAct against the state-of-the-art methods, in particular FT-ClipAct and FitAct.
It is observed that equipping DNNs with ProAct remarkably enhances the resilience of DNNs compared to the other state-of-the-art methods.

\begin{table*}[ht!]
\captionsetup{justification=centering}
\centering
\begin{tabular}{cc}
\multicolumn{2}{c}{
\begin{tikzpicture}
\begin{customlegend}[legend columns=5,legend style={text opacity =1,row sep=0pt, font=\fontsize{10}{8}\selectfont, column sep=1ex},
        legend entries={
                        \textsc{Ranger NW },
                        \textsc{Ranger LW },
                        \textsc{FT-ClipAct},
                        \textsc{FitAct},
                        \textsc{\textbf{ProAct}}
                        }]
        \addlegendimage{mark=*, mark size=2pt, cyan,draw=cyan}
        \addlegendimage{mark=*, mark size=2pt, blue,draw=blue}
        \addlegendimage{violet,mark=x,mark size=3pt}
        \addlegendimage{mark=square*, mark size=1.5pt, green, thick,draw=green}
        \addlegendimage{red,thick,mark=triangle*,mark size=2pt}
        \end{customlegend}
\end{tikzpicture}}
\\
\begin{tikzpicture}
\pgfplotsset{
  log x ticks with fixed point/.style={
      xticklabel={
        \pgfkeys{/pgf/fpu=true}
        \pgfmathparse{exp(\tick)}%
        \pgfmathprintnumber[fixed relative, precision=3]{\pgfmathresult}
        \pgfkeys{/pgf/fpu=false}
      }
  },
  log y ticks with fixed point/.style={
      yticklabel={
        \pgfkeys{/pgf/fpu=true}
        \pgfmathparse{exp(\tick)}%
        \pgfmathprintnumber[fixed relative, precision=3]{\pgfmathresult}
        \pgfkeys{/pgf/fpu=false}
      }
  }
}
  \begin{axis}[xmode=log,
        width=0.8\columnwidth,
        height=0.5\columnwidth,
        font=\footnotesize,
        scaled x ticks = false,
        scaled y ticks = false,
        xtick={0.0000001, 0.0000003, 0.000001,0.000003, 0.00001,0.00003}, 
        ytick={0.1,0.2,0.3,0.4,0.5,0.6,0.7,0.80},
        xticklabels = {\strut $10^{-7}$ , \strut $3\times10^{-7}$, \strut  $10^{-6}$,\strut $3\times10^{-6}$,$10^{-5}$,$3\times10^{-5}$},
        yticklabels = {\strut $10$,\strut $20$,\strut $30$,\strut $40$,\strut $50$, \strut $60$ ,\strut $70$, \strut $80$},
        ymin=0.1, ymax=0.82,
        xmin =0.0000001, xmax=0.00003,
        grid=major, 
        grid style={dashed,gray}, 
        ylabel near ticks,
        xlabel near ticks,
        xlabel= BER, 
        ylabel= Top-1 Accuracy (\%),
        ]
        \addplot [cyan,mark=*,mark size=2pt] table [x=BER, y=ranger-neuron, col sep=comma] {charts/accuray_AR_FI_alexnet_C10.csv};
        \addplot [blue,mark=*,mark size=2pt] table [x=BER, y=ranger-layer, col sep=comma] {charts/accuray_AR_FI_alexnet_C10.csv};
        \addplot [violet,mark=x,mark size=2pt] table [x=BER, y=ft-clipact, col sep=comma] {charts/accuray_AR_FI_alexnet_C10.csv};
        \addplot [green,mark=square*,mark size=1.5pt] table [x=BER, y=fitact-neuron, col sep=comma] {charts/accuray_AR_FI_alexnet_C10.csv};
        \addplot [red, thick,mark=triangle*,mark size=2pt] table [x=BER, y=protect, col sep=comma] {charts/accuray_AR_FI_alexnet_C10.csv};
      \end{axis}
    \end{tikzpicture}
  &
\begin{tikzpicture}
\pgfplotsset{
  log x ticks with fixed point/.style={
      xticklabel={
        \pgfkeys{/pgf/fpu=true}
        \pgfmathparse{exp(\tick)}%
        \pgfmathprintnumber[fixed relative, precision=3]{\pgfmathresult}
        \pgfkeys{/pgf/fpu=false}
      }
  },
  log y ticks with fixed point/.style={
      yticklabel={
        \pgfkeys{/pgf/fpu=true}
        \pgfmathparse{exp(\tick)}%
        \pgfmathprintnumber[fixed relative, precision=3]{\pgfmathresult}
        \pgfkeys{/pgf/fpu=false}
      }
  }
}
  \begin{axis}[xmode=log,
        width=0.8\columnwidth,
        height=0.5\columnwidth,
        font=\footnotesize,
        scaled x ticks = false,
        scaled y ticks = false,
        xtick={0.0000003, 0.000001, 0.000003}, 
        ytick={0.60, 0.65, 0.70, 0.75, 0.80},
        xticklabels = {  \strut $3\times10^{-7}$, \strut  $10^{-6}$,\strut $3\times10^{-6}$},
        yticklabels = {\strut $60$, \strut $65$, \strut $70$, \strut $75$, \strut $80$},
        ymin=0.6, ymax=0.82,
        xmin =0.00000028, xmax=0.0000032,
        grid=major, 
        grid style={dashed,gray}, 
        ylabel near ticks,
        xlabel near ticks,
        xlabel= BER, 
        ylabel= Top-1 Accuracy (\%),
        ]
        \addplot [violet,ultra thick,mark=x,mark size=3pt] table [x=BER, y=ft-clipact, col sep=comma] {charts/accuray_AR_FI_alexnet_C10.csv};
        \addplot [ green,ultra thick,mark=square*,mark size=1.5pt] table [x=BER, y=fitact-neuron, col sep=comma] {charts/accuray_AR_FI_alexnet_C10.csv};
        \addplot [red,ultra thick,mark=triangle*,mark size=2pt] table [x=BER, y=protect, col sep=comma] {charts/accuray_AR_FI_alexnet_C10.csv};
      \end{axis}
    \end{tikzpicture}

\\

\multicolumn{2}{c}{(a) AlexNet CIFAR-10}

\\
\begin{tikzpicture}
\pgfplotsset{
  log x ticks with fixed point/.style={
      xticklabel={
        \pgfkeys{/pgf/fpu=true}
        \pgfmathparse{exp(\tick)}%
        \pgfmathprintnumber[fixed relative, precision=3]{\pgfmathresult}
        \pgfkeys{/pgf/fpu=false}
      }
  },
  log y ticks with fixed point/.style={
      yticklabel={
        \pgfkeys{/pgf/fpu=true}
        \pgfmathparse{exp(\tick)}%
        \pgfmathprintnumber[fixed relative, precision=3]{\pgfmathresult}
        \pgfkeys{/pgf/fpu=false}
      }
  }
}
  \begin{axis}[xmode=log,
        width=0.8\columnwidth,
        height=0.5\columnwidth,
        font=\footnotesize,
        scaled x ticks = false,
        scaled y ticks = false,
        xtick={0.0000001, 0.0000003, 0.000001,0.000003,0.00001,0.00003}, 
        ytick={0.1,0.2,0.3,0.4,0.5,0.60,0.70,0.80,0.90},
        xticklabels = {\strut $10^{-7}$ , \strut $3\times10^{-7}$, \strut  $10^{-6}$,\strut $3\times10^{-6}$,\strut  $10^{-5}$,\strut $3\times10^{-5}$},
        yticklabels = {\strut $10$,\strut $20$,\strut $30$,\strut $40$,\strut $50$,\strut $60$,\strut $70$, \strut $80$ ,\strut $90$},
        ymin=0.1, ymax=0.9,
        xmin =0.0000001, xmax=0.00003,
        grid=major, 
        grid style={dashed,gray}, 
        ylabel near ticks,
        xlabel near ticks,
        xlabel= BER, 
        ylabel= Top-1 Accuracy (\%),
        ]
        \addplot [cyan,mark=*,mark size=2pt] table [x=BER, y=ranger-neuron, col sep=comma] {charts/accuray_AR_FI_vgg16_C10.csv};
        \addplot [blue,mark=*,mark size=2pt] table [x=BER, y=ranger-layer, col sep=comma] {charts/accuray_AR_FI_vgg16_C10.csv};
        \addplot [violet,mark=x,mark size=2pt] table [x=BER, y=ft-clipact, col sep=comma] {charts/accuray_AR_FI_vgg16_C10.csv};
        \addplot [green,mark=square*,mark size=1.5pt] table [x=BER, y=fitact-neuron, col sep=comma] {charts/accuray_AR_FI_vgg16_C10.csv};
        \addplot [red, thick,mark=triangle*,mark size=2pt] table [x=BER, y=protect, col sep=comma] {charts/accuray_AR_FI_vgg16_C10.csv};
      \end{axis}
    \end{tikzpicture}

    &

\begin{tikzpicture}
\pgfplotsset{
  log x ticks with fixed point/.style={
      xticklabel={
        \pgfkeys{/pgf/fpu=true}
        \pgfmathparse{exp(\tick)}%
        \pgfmathprintnumber[fixed relative, precision=3]{\pgfmathresult}
        \pgfkeys{/pgf/fpu=false}
      }
  },
  log y ticks with fixed point/.style={
      yticklabel={
        \pgfkeys{/pgf/fpu=true}
        \pgfmathparse{exp(\tick)}%
        \pgfmathprintnumber[fixed relative, precision=3]{\pgfmathresult}
        \pgfkeys{/pgf/fpu=false}
      }
  }
}
  \begin{axis}[xmode=log,
        width=0.8\columnwidth,
        height=0.5\columnwidth,
        font=\footnotesize,
        scaled x ticks = false,
        scaled y ticks = false,
        xtick={ 0.0000003, 0.000001,0.000003}, 
        ytick={0.865,0.87,0.875,0.88,0.885,0.89},
        xticklabels = { \strut $3\times10^{-7}$, \strut  $10^{-6}$,\strut $3\times10^{-6}$},
        yticklabels = {\strut $86.5$, \strut $87$, \strut $87.5$, \strut $88$, \strut $88.5$, \strut $89$},
        ymin=0.865, ymax=0.89,
        xmin =0.00000028, xmax=0.0000032,
        grid=major, 
        grid style={dashed,gray}, 
        ylabel near ticks,
        xlabel near ticks,
        xlabel= BER, 
        ylabel= Top-1 Accuracy (\%),
        ]
        \addplot [violet,ultra thick,mark=x,mark size=3pt] table [x=BER, y=ft-clipact, col sep=comma] {charts/accuray_AR_FI_vgg16_C10.csv};
        \addplot [green,ultra thick,mark=square*,mark size=1.5pt] table [x=BER, y=fitact-neuron, col sep=comma] {charts/accuray_AR_FI_vgg16_C10.csv};
        \addplot [red,ultra thick,mark=triangle*,mark size=2pt] table [x=BER, y=protect, col sep=comma] {charts/accuray_AR_FI_vgg16_C10.csv};
      \end{axis}
    \end{tikzpicture}

\\

\multicolumn{2}{c}{(b) VGG-16 CIFAR-10} 
\\

\begin{tikzpicture}
\pgfplotsset{
  log x ticks with fixed point/.style={
      xticklabel={
        \pgfkeys{/pgf/fpu=true}
        \pgfmathparse{exp(\tick)}%
        \pgfmathprintnumber[fixed relative, precision=3]{\pgfmathresult}
        \pgfkeys{/pgf/fpu=false}
      }
  },
  log y ticks with fixed point/.style={
      yticklabel={
        \pgfkeys{/pgf/fpu=true}
        \pgfmathparse{exp(\tick)}%
        \pgfmathprintnumber[fixed relative, precision=3]{\pgfmathresult}
        \pgfkeys{/pgf/fpu=false}
      }
  }
}
  \begin{axis}[xmode=log,
        width=0.8\columnwidth,
        height=0.5\columnwidth,
        font=\footnotesize,
        scaled x ticks = false,
        scaled y ticks = false,
        xtick={0.0000001, 0.0000003, 0.000001,0.000003,0.00001,0.00003}, 
        ytick={0.1,0.2,0.3,0.4,0.5,0.6,0.70,0.80,0.90},
        xticklabels = {\strut $10^{-7}$ , \strut $3\times10^{-7}$, \strut  $10^{-6}$,\strut $3\times10^{-6}$,\strut  $10^{-5}$,\strut $3\times10^{-5}$},
        yticklabels = {\strut $10$,\strut $20$,\strut $30$,\strut $40$,\strut $50$,\strut $60$,\strut $70$, \strut $80$ ,\strut $90$},
        ymin=0.1, ymax=0.92,
        xmin =0.0000001, xmax=0.00003,
        grid=major, 
        grid style={dashed,gray}, 
        ylabel near ticks,
        xlabel near ticks,
        xlabel= BER, 
        ylabel= Top-1 Accuracy (\%),
        ]
        \addplot [cyan,mark=*,mark size=2pt] table [x=BER, y=ranger-neuron, col sep=comma] {charts/accuray_AR_FI_resnet50_C10.csv};
        \addplot [blue,mark=*,mark size=2pt] table [x=BER, y=ranger-layer, col sep=comma] {charts/accuray_AR_FI_resnet50_C10.csv};
        \addplot [violet,mark=x,mark size=2pt] table [x=BER, y=ft-clipact, col sep=comma] {charts/accuray_AR_FI_resnet50_C10.csv};
        \addplot [green,mark=square*,mark size=1.5pt] table [x=BER, y=fitact-neuron, col sep=comma] {charts/accuray_AR_FI_resnet50_C10.csv};
        \addplot [red, thick,mark=triangle*,mark size=2pt] table [x=BER, y=protect, col sep=comma] {charts/accuray_AR_FI_resnet50_C10.csv};
      \end{axis}
    \end{tikzpicture}

&

\begin{tikzpicture}
\pgfplotsset{
  log x ticks with fixed point/.style={
      xticklabel={
        \pgfkeys{/pgf/fpu=true}
        \pgfmathparse{exp(\tick)}%
        \pgfmathprintnumber[fixed relative, precision=3]{\pgfmathresult}
        \pgfkeys{/pgf/fpu=false}
      }
  },
  log y ticks with fixed point/.style={
      yticklabel={
        \pgfkeys{/pgf/fpu=true}
        \pgfmathparse{exp(\tick)}%
        \pgfmathprintnumber[fixed relative, precision=3]{\pgfmathresult}
        \pgfkeys{/pgf/fpu=false}
      }
  }
}
  \begin{axis}[xmode=log,
        width=0.8\columnwidth,
        height=0.5\columnwidth,
        font=\footnotesize,
        scaled x ticks = false,
        scaled y ticks = false,
        xtick={0.0000001, 0.0000003, 0.000001}, 
        ytick={0.82, 0.84, 0.86, 0.88,0.90},
        xticklabels = {\strut $10^{-7}$ , \strut $3\times10^{-7}$, \strut  $10^{-6}$,\strut $3\times10^{-6}$},
        yticklabels = {\strut $82$, \strut $84$, \strut $86$, \strut $88$, \strut $90$},
        ymin=0.82, ymax=0.91,
        xmin =0.000000095, xmax=0.0000011,
        grid=major, 
        grid style={dashed,gray}, 
        ylabel near ticks,
        xlabel near ticks,
        xlabel= BER, 
        ylabel= Top-1 Accuracy (\%),
        ]
        \addplot [violet,ultra thick,mark=x,mark size=3pt] table [x=BER, y=ft-clipact, col sep=comma] {charts/accuray_AR_FI_resnet50_C10.csv};
        \addplot [green,ultra thick,mark=square*,mark size=1.5pt] table [x=BER, y=fitact-neuron, col sep=comma] {charts/accuray_AR_FI_resnet50_C10.csv};
        \addplot [red,ultra thick,mark=triangle*,mark size=2pt] table [x=BER, y=protect, col sep=comma] {charts/accuray_AR_FI_resnet50_C10.csv};
      \end{axis}
    \end{tikzpicture}

  \\

\multicolumn{2}{c}{(e) ResNet-50 CIFAR-10}

\end{tabular}
\captionof{figure}{Top-1 accuracy comparison of DNNs using ProAct with Ranger neuron-wise , Ranger layer-wise , FT-ClipAct, and FitAct methods under fault injection.}
\label{fig:experiments:acc:cifar10}
\end{table*}

\begin{table}[ht]
\centering
\caption{Average $L_2$ distance of different DNNs and mitigation techniques, layer-wise FI, BER = $1E-4$.}
\label{tab:l2-dist}
\begin{tabular}{cccc}
\hline
\multirow{3}{*}{Method} & 
\multicolumn{3}{c}{\textbf{$L_2$ Distance}}
\\\cline{2-4}
&{\textbf{AlexNet}} & {\textbf{VGG-16}} & {\textbf{ResNet-50}} \\
\hline
FitAct   & 65.6  & 80.3 & 93.7 \\
\textbf{ProAct}  & \textbf{45.9} &  \textbf{72.4}  & \textbf{86.5} \\
\hline
\end{tabular}
\end{table}
\begin{figure}
    \centering
    \includegraphics[width=\columnwidth, height=\columnwidth]{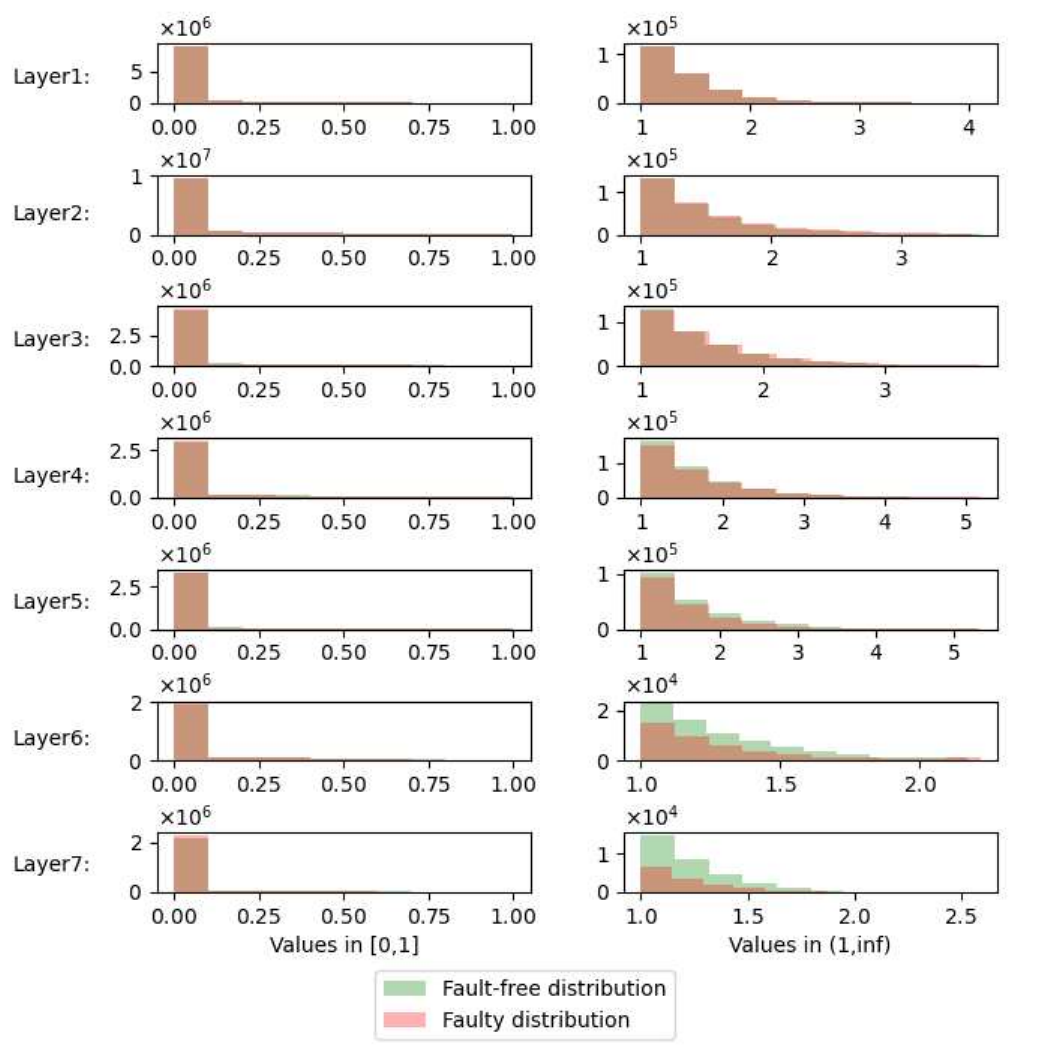}
    \caption{The distribution of output activation values for the AlexNet model on the CIFAR-10 dataset after applying the ProAct algorithm to find threshold parameters.  }
    \label{fig:ProAct_Dist}
\end{figure}

\begin{table*}[ht!]
\captionsetup{justification=centering}
\centering
\begin{tabular}{cc}
\multicolumn{2}{c}{
\begin{tikzpicture}
\begin{customlegend}[legend columns=5,legend style={text opacity =1,row sep=0pt, font=\fontsize{12}{8}\selectfont, column sep=1ex},
        legend entries={
                        \textsc{Ranger NW },
                        \textsc{Ranger LW },
                        \textsc{FT-ClipAct},
                        \textsc{FitAct},
                        \textsc{\textbf{ProAct}}
                        }]
        \addlegendimage{mark=*, mark size=2pt, cyan,draw=cyan}
        \addlegendimage{mark=*, mark size=2pt, blue,draw=blue}
        \addlegendimage{violet,mark=x,mark size=3pt}
        \addlegendimage{mark=square*, mark size=1.5pt, green, thick,draw=green}
        \addlegendimage{red,thick,mark=triangle*,mark size=2pt}
        \end{customlegend}
\end{tikzpicture}}
\\
\begin{tikzpicture}
\pgfplotsset{
  log x ticks with fixed point/.style={
      xticklabel={
        \pgfkeys{/pgf/fpu=true}
        \pgfmathparse{exp(\tick)}%
        \pgfmathprintnumber[fixed relative, precision=3]{\pgfmathresult}
        \pgfkeys{/pgf/fpu=false}
      }
  },
  log y ticks with fixed point/.style={
      yticklabel={
        \pgfkeys{/pgf/fpu=true}
        \pgfmathparse{exp(\tick)}%
        \pgfmathprintnumber[fixed relative, precision=3]{\pgfmathresult}
        \pgfkeys{/pgf/fpu=false}
      }
  }
}
  \begin{axis}[xmode=log,
        width=0.8\columnwidth,
        height=0.5\columnwidth,
        font=\footnotesize,
        scaled x ticks = false,
        scaled y ticks = false,
        xtick={0.0000001, 0.0000003, 0.000001,0.000003,0.00001,0.00003}, 
        ytick={0.0,0.10,0.20,0.30,0.4,0.50,0.54},
        xticklabels = {\strut $10^{-7}$ , \strut $3\times10^{-7}$, \strut  $10^{-6}$,\strut  $3\times10^{-6}$,\strut  $10^{-5}$,\strut  $3\times10^{-5}$},
       yticklabels = {\strut $0$,\strut $10$,\strut $20$, \strut $30$ ,\strut $40$,\strut $50$,\strut $54$},
        ymin=0.0, ymax=0.55,
        xmin =0.0000001, xmax=0.00003,
        grid=major, 
        grid style={dashed,gray}, 
        ylabel near ticks,
        xlabel near ticks,
        xlabel= BER, 
        ylabel= Top-1 Accuracy (\%),
        ]
        \addplot [cyan,mark=*,mark size=2pt] table [x=BER, y=ranger-neuron, col sep=comma] {charts/accuray_AR_FI_alexnet_C100.csv};
        \addplot [blue,mark=*,mark size=2pt] table [x=BER, y=ranger-layer, col sep=comma] {charts/accuray_AR_FI_alexnet_C100.csv};
        \addplot [violet,mark=x,mark size=2pt] table [x=BER, y=ft-clipact, col sep=comma] {charts/accuray_AR_FI_alexnet_C100.csv};
        \addplot [green,mark=square*,mark size=2pt] table [x=BER, y=fitact-neuron, col sep=comma] {charts/accuray_AR_FI_alexnet_C100.csv};
        \addplot [red,thick,mark=triangle*,mark size=2pt] table [x=BER, y=protect, col sep=comma] {charts/accuray_AR_FI_alexnet_C100.csv};
      \end{axis}
    \end{tikzpicture}
    &
\begin{tikzpicture}
\pgfplotsset{
  log x ticks with fixed point/.style={
      xticklabel={
        \pgfkeys{/pgf/fpu=true}
        \pgfmathparse{exp(\tick)}%
        \pgfmathprintnumber[fixed relative, precision=3]{\pgfmathresult}
        \pgfkeys{/pgf/fpu=false}
      }
  },
  log y ticks with fixed point/.style={
      yticklabel={
        \pgfkeys{/pgf/fpu=true}
        \pgfmathparse{exp(\tick)}%
        \pgfmathprintnumber[fixed relative, precision=3]{\pgfmathresult}
        \pgfkeys{/pgf/fpu=false}
      }
  }
}
  \begin{axis}[xmode=log,
        width=0.8\columnwidth,
        height=0.5\columnwidth,
        font=\footnotesize,
        scaled x ticks = false,
        scaled y ticks = false,
        xtick={0.0000001, 0.0000003}, 
        ytick={0.47,0.485,0.50,0.515,0.53},
        xticklabels = {\strut $10^{-7}$ , \strut $3\times10^{-7}$},
       yticklabels = {\strut $47$,\strut $48.5$,\strut $50$,\strut $51.5$,\strut $53$},
        ymin=0.465, ymax=0.535,
        xmin =0.0000001, xmax=0.00000032,
        grid=major, 
        grid style={dashed,gray}, 
        ylabel near ticks,
        xlabel near ticks,
        xlabel= BER, 
        ylabel= Top-1 Accuracy (\%),
        ]
        \addplot [violet,ultra thick,mark=x,mark size=3pt] table [x=BER, y=ft-clipact, col sep=comma] {charts/accuray_AR_FI_alexnet_C100.csv};
        \addplot [green,ultra thick,mark=square*,mark size=1.5pt] table [x=BER, y=fitact-neuron, col sep=comma] {charts/accuray_AR_FI_alexnet_C100.csv};
        \addplot [red,ultra thick,mark=triangle*,mark size=2pt] table [x=BER, y=protect, col sep=comma] {charts/accuray_AR_FI_alexnet_C100.csv};
      \end{axis}
    \end{tikzpicture}

\\

\multicolumn{2}{c}{(a) AlexNet CIFAR-100}
\\

\begin{tikzpicture}
\pgfplotsset{
  log x ticks with fixed point/.style={
      xticklabel={
        \pgfkeys{/pgf/fpu=true}
        \pgfmathparse{exp(\tick)}%
        \pgfmathprintnumber[fixed relative, precision=3]{\pgfmathresult}
        \pgfkeys{/pgf/fpu=false}
      }
  },
  log y ticks with fixed point/.style={
      yticklabel={
        \pgfkeys{/pgf/fpu=true}
        \pgfmathparse{exp(\tick)}%
        \pgfmathprintnumber[fixed relative, precision=3]{\pgfmathresult}
        \pgfkeys{/pgf/fpu=false}
      }
  }
}
  \begin{axis}[xmode=log,
        width=0.8\columnwidth,
        height=0.5\columnwidth,
        font=\footnotesize,
        scaled x ticks = false,
        scaled y ticks = false,
        xtick={0.0000001, 0.0000003, 0.000001,0.000003,0.00001,0.00003}, 
        ytick={0.1,0.2,0.30,0.40,0.5,0.6},
        xticklabels = {\strut $10^{-7}$ , \strut $3\times10^{-7}$, \strut  $10^{-6}$,\strut $3\times10^{-6}$, \strut  $10^{-5}$,\strut $3\times10^{-5}$, },
       yticklabels = {\strut $10$,\strut $20$,\strut $30$, \strut $40$ ,\strut $50$,\strut $60$},
        ymin=0.0, ymax=0.64,
        xmin =0.0000001, xmax=0.00003,
        grid=major, 
        grid style={dashed,gray}, 
        ylabel near ticks,
        xlabel near ticks,
        xlabel= BER, 
        ylabel= Top-1 Accuracy (\%),
        ]
        \addplot [cyan,mark=*,mark size=2pt] table [x=BER, y=ranger-neuron, col sep=comma] {charts/accuray_AR_FI_vgg16_C100.csv};
        \addplot [blue,mark=*,mark size=2pt] table [x=BER, y=ranger-layer, col sep=comma] {charts/accuray_AR_FI_vgg16_C100.csv};
        \addplot [violet,mark=x,mark size=2pt] table [x=BER, y=ft-clipact, col sep=comma] {charts/accuray_AR_FI_vgg16_C100.csv};
        \addplot [green,mark=square*,mark size=2pt] table [x=BER, y=fitact-neuron, col sep=comma] {charts/accuray_AR_FI_vgg16_C100.csv};
        \addplot [red, thick,mark=triangle*,mark size=2pt] table [x=BER, y=protect, col sep=comma] {charts/accuray_AR_FI_vgg16_C100.csv};
      \end{axis}
    \end{tikzpicture}

&
\begin{tikzpicture}
\pgfplotsset{
  log x ticks with fixed point/.style={
      xticklabel={
        \pgfkeys{/pgf/fpu=true}
        \pgfmathparse{exp(\tick)}%
        \pgfmathprintnumber[fixed relative, precision=3]{\pgfmathresult}
        \pgfkeys{/pgf/fpu=false}
      }
  },
  log y ticks with fixed point/.style={
      yticklabel={
        \pgfkeys{/pgf/fpu=true}
        \pgfmathparse{exp(\tick)}%
        \pgfmathprintnumber[fixed relative, precision=3]{\pgfmathresult}
        \pgfkeys{/pgf/fpu=false}
      }
  }
}
  \begin{axis}[xmode=log,
        width=0.8\columnwidth,
        height=0.5\columnwidth,
        font=\footnotesize,
        scaled x ticks = false,
        scaled y ticks = false,
        xtick={0.0000001,0.0000003}, 
        ytick={0.45,0.50,0.55,0.60},
        xticklabels = {\strut $10^{-7}$,\strut $3\times10^{-7}$},
       yticklabels = {\strut $45$,\strut $50$, \strut $55$ ,\strut $60$},
        ymin=0.4, ymax=0.63,
        xmin =0.0000001, xmax=0.00000032,
        grid=major, 
        grid style={dashed,gray}, 
        ylabel near ticks,
        xlabel near ticks,
        xlabel= BER, 
        ylabel= Top-1 Accuracy (\%),
        ]
        \addplot [violet,ultra thick,mark=x,mark size=3pt] table [x=BER, y=ft-clipact, col sep=comma] {charts/accuray_AR_FI_vgg16_C100.csv};
        \addplot [green,ultra thick,mark=square*,mark size=1.5pt] table [x=BER, y=fitact-neuron, col sep=comma] {charts/accuray_AR_FI_vgg16_C100.csv};
        \addplot [red,ultra thick,mark=triangle*,mark size=2pt] table [x=BER, y=protect, col sep=comma] {charts/accuray_AR_FI_vgg16_C100.csv};
      \end{axis}
    \end{tikzpicture}

\\

\multicolumn{2}{c}{(b) VGG-16 CIFAR-100}
\\

\begin{tikzpicture}
\pgfplotsset{
  log x ticks with fixed point/.style={
      xticklabel={
        \pgfkeys{/pgf/fpu=true}
        \pgfmathparse{exp(\tick)}%
        \pgfmathprintnumber[fixed relative, precision=3]{\pgfmathresult}
        \pgfkeys{/pgf/fpu=false}
      }
  },
  log y ticks with fixed point/.style={
      yticklabel={
        \pgfkeys{/pgf/fpu=true}
        \pgfmathparse{exp(\tick)}%
        \pgfmathprintnumber[fixed relative, precision=3]{\pgfmathresult}
        \pgfkeys{/pgf/fpu=false}
      }
  }
}
  \begin{axis}[xmode=log,
        width=0.8\columnwidth,
        height=0.5\columnwidth,
        font=\footnotesize,
        scaled x ticks = false,
        scaled y ticks = false,
        xtick={0.0000001, 0.0000003, 0.000001,0.000003,0.00001,0.00003}, 
        ytick={0.05,0.15,0.25,0.35,0.45,0.55,0.65,0.75},
        xticklabels = {\strut $10^{-7}$ , \strut $3\times10^{-7}$, \strut  $10^{-6}$,\strut $3\times10^{-6}$,\strut  $10^{-5}$,\strut $3\times10^{-5}$},
       yticklabels = {\strut $5$,\strut $15$,\strut $25$,\strut $35$,\strut $45$, \strut $55$ ,\strut $65$,\strut $75$},
        ymin=0.0, ymax=0.75,
        xmin =0.0000001, xmax=0.00003,
        grid=major, 
        grid style={dashed,gray}, 
        ylabel near ticks,
        xlabel near ticks,
        xlabel= BER, 
        ylabel= Top-1 Accuracy (\%),
        ]
        \addplot [cyan,mark=*,mark size=2pt] table [x=BER, y=ranger-neuron, col sep=comma] {charts/accuray_AR_FI_resnet50_C100.csv};
        \addplot [blue,mark=*,mark size=2pt] table [x=BER, y=ranger-layer, col sep=comma] {charts/accuray_AR_FI_resnet50_C100.csv};
        \addplot [violet,mark=x,mark size=2pt] table [x=BER, y=ft-clipact, col sep=comma] {charts/accuray_AR_FI_resnet50_C100.csv};
        \addplot [green,mark=square*,mark size=2pt] table [x=BER, y=fitact-neuron, col sep=comma] {charts/accuray_AR_FI_resnet50_C100.csv};
        \addplot [red, thick,mark=triangle*,mark size=2pt] table [x=BER, y=protect, col sep=comma] {charts/accuray_AR_FI_resnet50_C100.csv};
      \end{axis}
    \end{tikzpicture}
&
     
\begin{tikzpicture}
\pgfplotsset{
  log x ticks with fixed point/.style={
      xticklabel={
        \pgfkeys{/pgf/fpu=true}
        \pgfmathparse{exp(\tick)}%
        \pgfmathprintnumber[fixed relative, precision=3]{\pgfmathresult}
        \pgfkeys{/pgf/fpu=false}
      }
  },
  log y ticks with fixed point/.style={
      yticklabel={
        \pgfkeys{/pgf/fpu=true}
        \pgfmathparse{exp(\tick)}%
        \pgfmathprintnumber[fixed relative, precision=3]{\pgfmathresult}
        \pgfkeys{/pgf/fpu=false}
      }
  }
}
  \begin{axis}[xmode=log,
        width=0.8\columnwidth,
        height=0.5\columnwidth,
        font=\footnotesize,
        scaled x ticks = false,
        scaled y ticks = false,
        xtick={0.0000001, 0.0000003}, 
        ytick={0.6,0.65,0.7,0.75},
        xticklabels = {\strut $10^{-7}$ , \strut $3\times10^{-7}$},
       yticklabels = {\strut $60$,\strut $65$,\strut $70$,\strut $75$},
        ymin=0.6, ymax=0.75,
        xmin =0.0000001, xmax=0.00000033,
        grid=major, 
        grid style={dashed,gray}, 
        ylabel near ticks,
        xlabel near ticks,
        xlabel= BER, 
        ylabel= Top-1 Accuracy (\%),
        ]
        \addplot [violet,ultra thick,mark=x,mark size=3pt] table [x=BER, y=ft-clipact, col sep=comma] {charts/accuray_AR_FI_resnet50_C100.csv};
        \addplot [green,ultra thick,mark=square*,mark size=1.5pt] table [x=BER, y=fitact-neuron, col sep=comma] {charts/accuray_AR_FI_resnet50_C100.csv};
        \addplot [red,ultra thick,mark=triangle*,mark size=2pt] table [x=BER, y=protect, col sep=comma] {charts/accuray_AR_FI_resnet50_C100.csv};
      \end{axis}
    \end{tikzpicture}

    \\

\multicolumn{2}{c}{(c) ResNet-50 CIFAR-100}
\end{tabular}

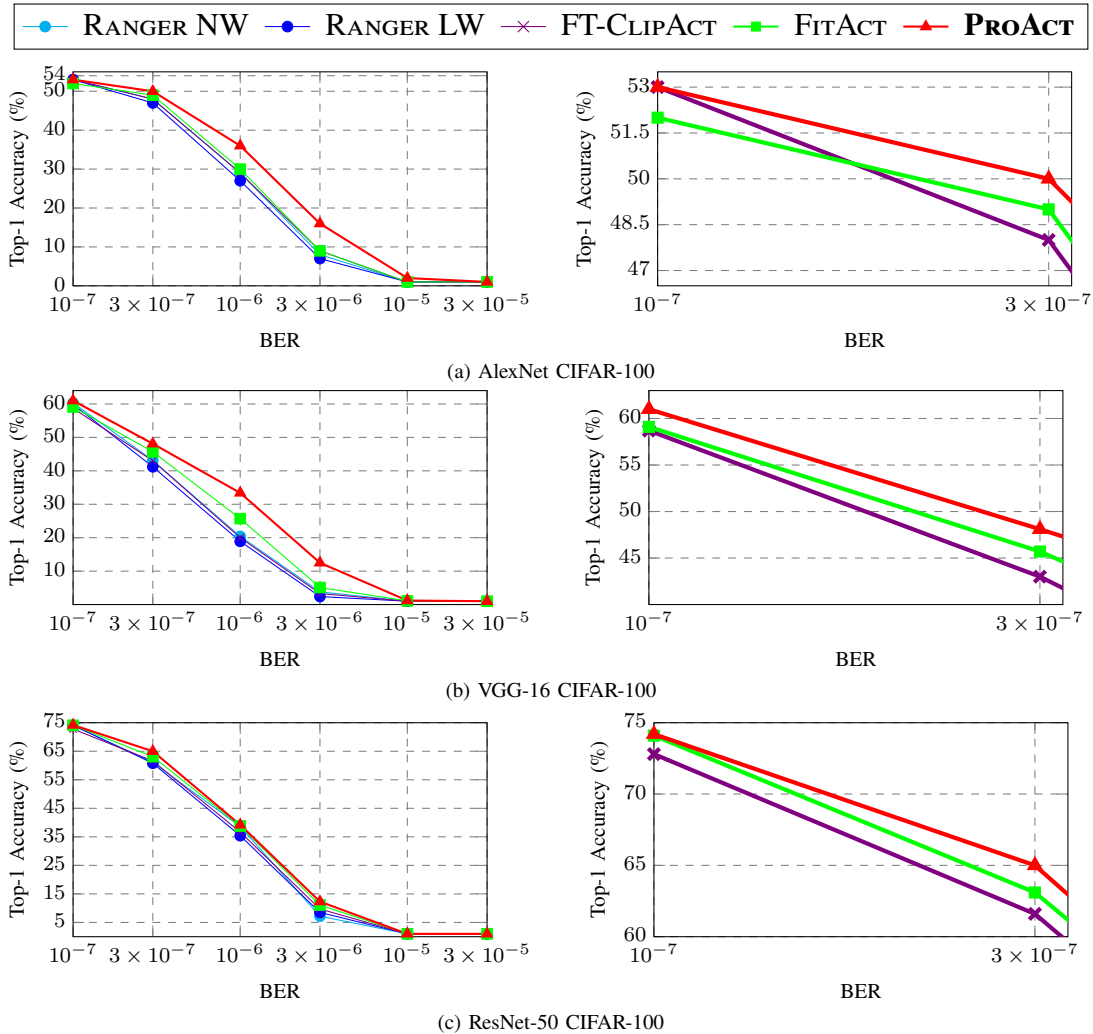
\captionof{figure}{Top-1 accuracy comparison of DNNs using ProAct with Ranger neuron-wise, Ranger layer-wise, FT-ClipAct, and FitAct methods under fault injection.}
\label{fig:experiments:acc}
\end{table*}

Regarding Fig.~\ref{fig:experiments:acc:cifar10} and Fig.~\ref{fig:experiments:acc}, at all BERs, the accuracy of DNNs with ProAct is higher than the DNNs with other activation restriction methods.
As it is observed, Ranger provides the least resilient DNNs.
According to the results, although FitAct provides better resilience than FT-ClipAct, it introduces a remarkable memory overhead orders of magnitude more than FT-ClipAct.
Whereas ProAct achieves a higher resilience than all existing methods with negligible overhead. 

Moreover, it is observed that in model-wise FI experiments, all activation restriction methods can effectively improve the resilience of DNNs compared to unprotected DNNs. However, they fail to provide highly resilient DNNs at high BERs.
When faulty weights are spread throughout a DNN, several neurons in various layers are affected. 
Consequently, in a high BER, the values of affected neurons are restricted by their activation functions and make numerous erroneous activations propagate to the DNN output resulting in a considerable accuracy drop. 
However, ProAct surpasses the other activation restriction techniques in terms of providing superior accuracy for DNNs in model-wise FI.

Table~\ref{tab:acc-drop-ber} summarizes the results for accuracy drop of experimented DNNs with respect to their own baseline accuracy in Table \ref{table:params-count}, hardened by FT-ClipAct, FitAct and ProAct, at the BERs where the accuracy drop of ProActed DNNs for CIFAR-10 is less than 5\%, and for CIFAR-100 is less than 10\%. This comparison implicitly includes the accuracy drop due to activation restriction methods exhibiting the overall benefit of ProAct. 
According to the results, it is observed that ProAct reduces the accuracy drop of DNNs from 1.36x up to 6.4x compared to FT-ClipAct and from 1.07x up to 1.72x compared to FitAct.

\subsubsection{\textbf{Activation Distribution in ProActed DNNs}}
As discussed in Section \ref{sec:Motivation} and illustrated in Fig.~\ref{fig:enter-label}, there is a significant difference in the distribution of activation values between the fault-free and faulty models, particularly in the last layer.
ProAct addresses this issue by reducing the gap between these distributions through the identification of more effective thresholds for the hybrid activation functions, using a progressive training mechanism.

Fig.~\ref{fig:ProAct_Dist} presents the distribution of fault-free and faulty values after applying ProAct to the AlexNet model on the CIFAR-10 dataset. 
The figure demonstrates that the distributions are more similar, indicating that ProAct generates a model that closely resembles the fault-free model.
Specifically, incorporating Knowledge Distillation (KD) in the training process helps identify threshold values that maintain the similarity between the distributions of the fault-free model and the faulty model.

To numerically evaluate the similarity of the activation values between the clipped model and the fault-free model, we compute the average $L_2$ distance metric for all the layers of the models as:
\begin{equation}
    L_2  =  \frac{1}{N \times L}\sum_{i=1}^{L} \sum_{j=1}^{N} | z_{ij} - f_{ij}|_2^2
\end{equation}
where $L$ and $N$ show the number of DNN's layers and neurons in a layer, respectively. Also, $z$ and $f$ indicate the activation value for faulty and fault-free models, respectively.

Table~\ref{tab:l2-dist} compares the similarity metric for ProAct and FitAct as the state-of-the-art optimization based method. 
As observed, the $L_2$ distance of activations for all models is decreased by the ProAct, meaning it produces activations much closer to the fault-free network than FitAct. Compared to FitAct, ProAct improves the L2 distance by $30\%$, $9.83\%$, and $7.68\%$  on AlexNet, VGG-16 and ResNet-50, respectively.

\section{Conclusion} \label{conclusion}
In this work, we introduce ProAct, a progressive training method for determining threshold values in a novel hybrid clipped ReLU (HyReLU) activation function aimed at enhancing the resilience of DNNs. 
We demonstrate that existing optimization techniques are computationally intensive and frequently fail to find optimal threshold values.
Additionally, neuron-wise clipping methods such as FitAct are shown to incur substantial memory costs.
To address these issues, we develop a hybrid clipped ReLU activation function that reduces memory overhead by applying neuron-wise clipping solely in the final layer and layer-wise clipping in the preceding layers.
Following this, we proposed a progressive training strategy utilizing knowledge distillation to train the threshold parameters layer by layer, effectively identifying suitable threshold values for the HyReLU.
Our experimental results indicate that ProAct significantly improves the resilience of DNNs, with enhancements of up to $6.4x$ in high bit error rates.
Furthermore, our approach dramatically reduces memory overhead, achieving reductions of $10.5\times$ to $134.28\times$ compared to the leading neuron-wise activation restriction methods. Furthermore, we have published all source codes in Python, enabling researchers to present more effective approaches in this area. 

\section*{}
\bibliographystyle{IEEEtran}
\bibliography{references.bib}

\begin{IEEEbiography}[{\includegraphics[width=1in,height=1.25in,clip,keepaspectratio]{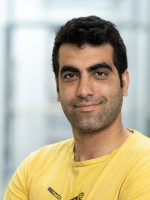}}]%
{
Seyedhamidreza Mousavi} is currently pursuing the PhD degree in computer science and engineering at the School of Innovation, Design, and Engineering, Mälardalen University, Västerås, Sweden. He is a member of AutoDeep and FASTER-AI projects. His research is focused on designing deep neural network architectures that are both high-performing and compact, while also ensuring their safety using reliable and robust neural architecture search techniques.
\end{IEEEbiography}

\begin{IEEEbiography}[{\includegraphics[width=1in,height=1.25in,clip,keepaspectratio]{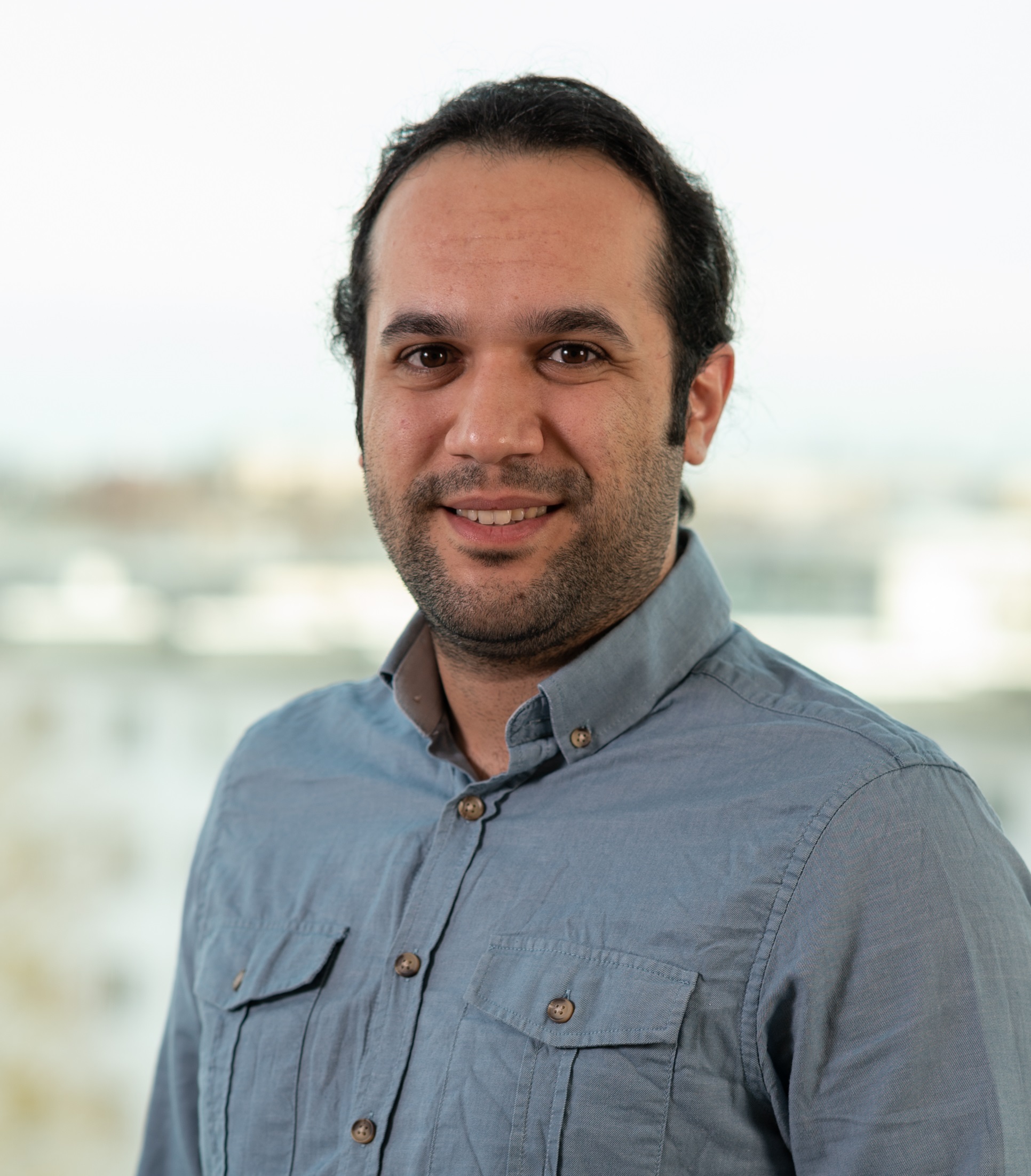}}]%
{
Mohammad Hasan Ahmadilivani} is a PhD student in the Computer Systems Department at Tallinn University of Technology (Taltech), Estonia. He earned his MSc in Computer Architecture Systems from the University of Tehran, Iran, in 2020. His research focuses on developing analytical methods to measure and enhance the hardware reliability of deep neural networks (DNNs). His research interests include exploiting DNNs in safety-critical applications, robust computer vision, and efficient and reliable DNN accelerator design.
\end{IEEEbiography}

\begin{IEEEbiography}[{\includegraphics[width=1in,height=1.25in,clip,keepaspectratio]{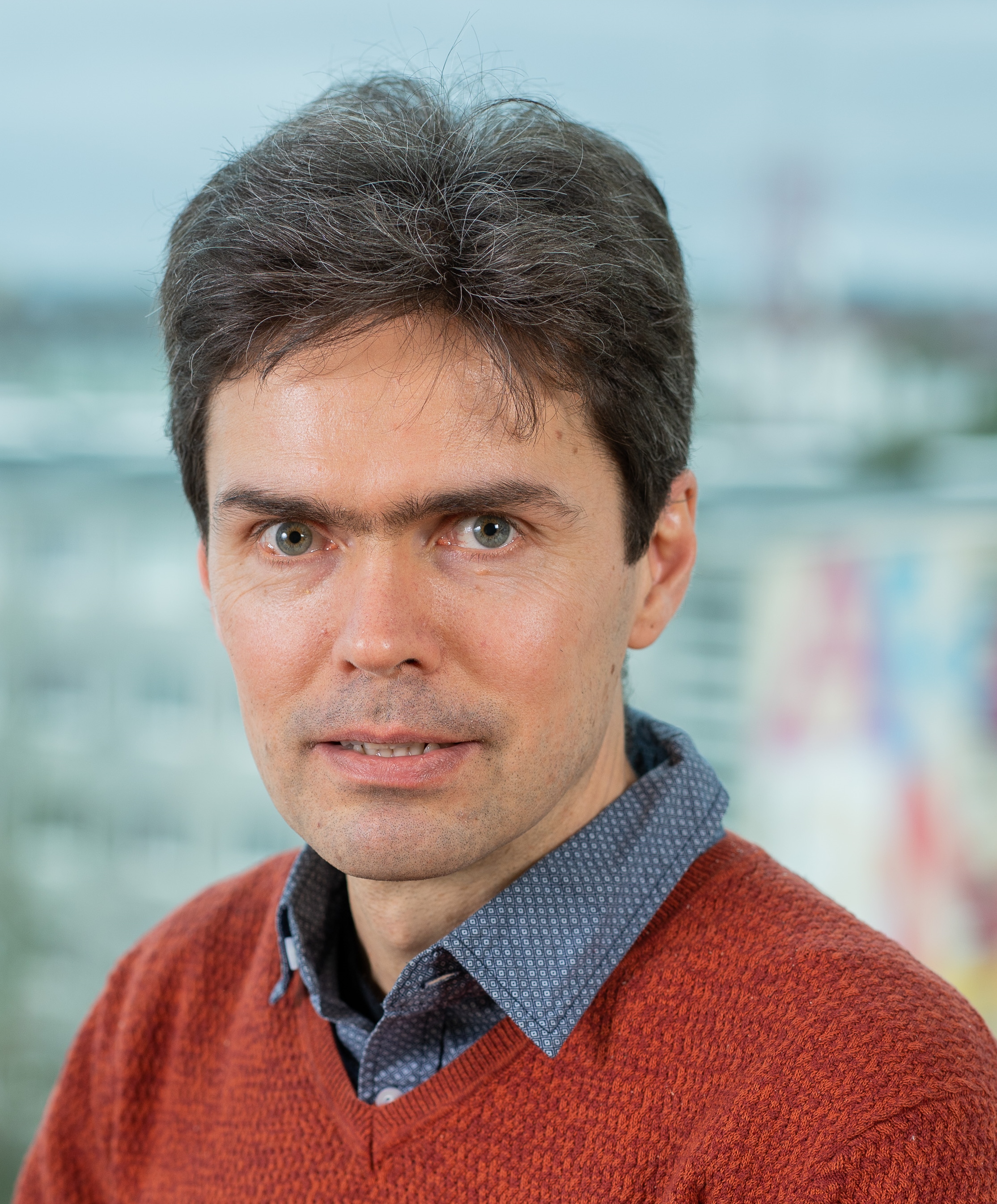}}]%
{
Jaan Raik} is a Full Professor at the Department of Computer Systems and Head of the Center for Dependable Computing Systems at the Tallinn University of Technology (Taltech), Estonia. He received his M.Sc. and Ph.D. degrees from Taltech in 1997 and in 2001, respectively. His research interests cover a wide area in electrical engineering and computer science domains including reliability of deep learning, hardware test, functional verification, fault-tolerance and security as well as emerging computer architectures. He has co-authored more than 400 scientific publications. He is a member of IEEE Computer Society, HiPEAC and of Steering/Program Committees of numerous conferences within his field. He served as the General Chair to IEEE European Test Symposium ’25, ’20, IFIP/IEEE VLSI-SoC ’16, DDECS ’12), Vice General Chair IEEE European Test Symposium ’24, DDECS ’13 and Program Co-Chair DDECS ’23, ’15, CDN-Live ’16 conferences. He was awarded the Global Digital Governance Fellowship at Stanford (2022), HiPEAC Paper Award (2020), the Order of the White Star 4th class medal by the President of Estonia (2016) and Estonian Academy of Science’s Bernhard Schmidt Award for innovation (2007).
\end{IEEEbiography}
\begin{IEEEbiography}[{\includegraphics[width=1in,height=1.25in,clip,keepaspectratio]{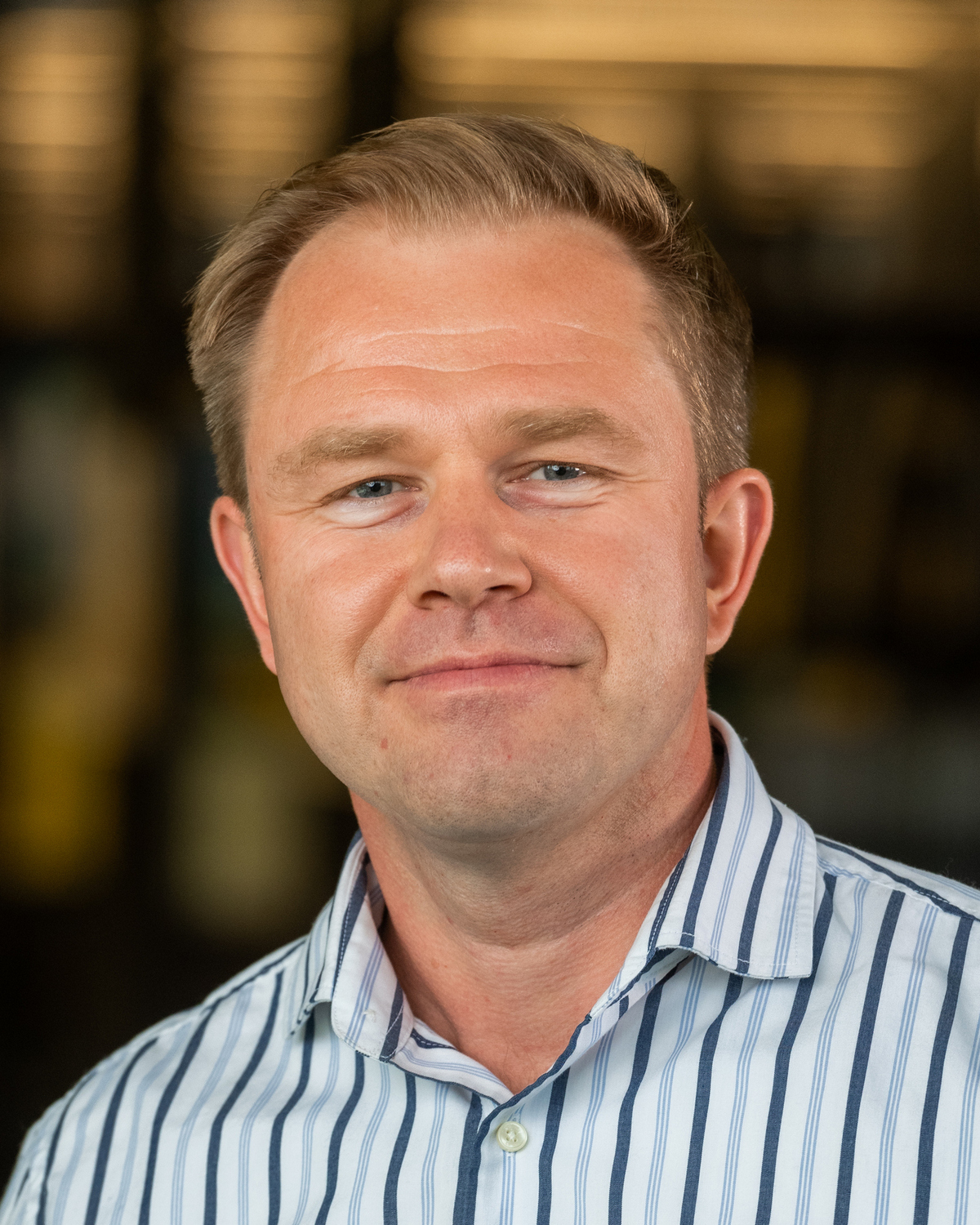}}]%
{
Maksim Jenihhin} is a tenured associate professor of computing systems reliability and Head of the research group Trustworthy and Efficient Computing Hardware (TECH) at the Tallinn University of Technology, Estonia. He received his PhD degree in Computer Engineering from the same university in 2008. His research interests include methodologies and EDA tools for hardware design, verification and debug, and security, as well as nanoelectronics reliability and manufacturing test topics. He has published more than 150 research papers, supervised several PhD students and postdocs and served on executive and program committees for numerous IEEE conferences (DATE, ETS, DDECS, LATS, NORCAS, etc.). Prof. Jenihhin coordinates European collaborative research projects HORIZON MSCA DN “TIRAMISU” (2024), HORIZON TWINN “TAICHIP” (2024) and national ones about energy efficiency and reliability of edge-AI chips and cross-layer self-health awareness of autonomous systems.
\end{IEEEbiography}

\begin{IEEEbiography}[{\includegraphics[width=1in,height=1.25in,clip,keepaspectratio]{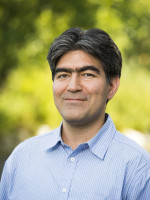}}]%
{
Masoud Daneshtalab} Masoud Daneshtalab is currently a full Prof. at Mälardalen University in Sweden, Adj. Prof. at TalTech in Estonia and leads the Heterogeneous System research group. He is on the Euromicro board of directors, an editor of the MICPRO journal, and has published over 200 refereed papers. His research interests include HW/SW/Algorithm co-design, dependability and deep learning acceleration.
\end{IEEEbiography}
\end{document}